\pdfoutput=1
\documentclass[11pt]{article}

\usepackage[]{ACL2023}

\usepackage{times}
\usepackage{latexsym}
\usepackage{ctable}

\usepackage[T1]{fontenc}
\usepackage[utf8]{inputenc}

\usepackage{microtype}

\usepackage{inconsolata}

\usepackage{array}
\usepackage{tabularx}
\usepackage{booktabs} 
\usepackage{caption}
\usepackage{siunitx}

\usepackage{cleveref}

\usepackage{graphicx}

\usepackage{color}
\usepackage{xcolor}
\newcommand{\sm}[1]{{\color{black} #1}}
\newcommand{\sw}[1]{{\color{black} #1}}

\title{The Language of Interoception: Examining Embodiment and Emotion Through a Corpus of Body Part Mentions} 

\author{
Sophie Wu \hspace*{2cm}\\
McGill University, Canada \hspace*{2cm}\\
\texttt{sophie.wu@mail.mcgill.ca} \hspace{2cm}
\And
\hspace{2cm} Jan Philip Wahle \\
\hspace{2cm} University of Göttingen, Germany \\
\hspace{2cm} \texttt{wahle@uni-goettingen.de}
\AND
Saif M. Mohammad \\
National Research Council Canada \\
\texttt{saif.mohammad@nrc-cnrc.gc.ca}
}

\begin{document}
\maketitle
\begin{abstract}
This paper is the first investigation of the connection between emotion, embodiment, and everyday language in a large sample of natural language data. First, we created corpora of body part mentions (BPMs) in online English text (blog posts and tweets). These include a subset featuring human annotations for the emotions of the person whose body part is mentioned in the text. Next, we show that BPMs are common in personal narratives and tweets ($\sim$5\% to 10\% of posts include BPMs) and that their usage patterns vary markedly by time and %geographic 
location. Using word--emotion association lexicons and our annotated data, we show that text containing BPMs tends to be more emotionally charged
\sm{than text without any BPMs.}
%even when the BPM is not explicitly used to describe a physical reaction to the emotion in the text. 
Finally, we %discover 
\sm{show} a strong and statistically significant correlation between body-related language and a variety of %poorer 
\sm{negative} health outcomes. In sum, we argue that investigating the role of body-part related words in language can open up valuable avenues of future research at the intersection of NLP, the affective sciences, and the study of human wellbeing.

\end{abstract}

\section{Introduction}

Embodied cognition---the theory that human cognition is rooted in bodily experiences---has gained significant traction across cognitive science, psychology, linguistics, and philosophy. This framework suggests that our bodily experiences shape not only how we interact with our physical environment, but also how we process, represent, and share abstract concepts. A wide range of disciplines have demonstrated that the intersection of cognition and language is deeply intertwined with sensorimotor experiences. For example, a growing pool of research suggests that: individuals learn new concepts better when they can use their bodies to simulate the concepts \cite{cook2008gesturing, johnson2016effects}; language processing involves mental simulation of physical actions \cite{pulvermuller2005brain, glenberg2002grounding}; and emotions emerge from interpretations of physiological signals through a process known as interoception \cite{craig2002you, barrett2017theory}.

% While it is well established that embodiment holds a key role in human cognition, [Already said that so commenting it]
% An important open question is the extent to which our embodied experiences are encoded in and reflected through everyday language.
\sm{In this paper, we ask the question:
\textit{to what extent are our embodied experiences encoded in, and reflected through, everyday language?}}
% NLP corpora and techniques offer a unique opportunity to systematically investigate this relationship at scale. 
To this end, we %introduce 
\sm{use} the concept of \textbf{Body Part Mentions (BPMs)}, which we define as \textit{instances of language where words 
% associated with 
\sm{referring to}
parts of the body are used}.\footnote{\sm{Some linguistics paper use the term \textit{somatic reference} for such mentions, and the term \textit{somatic unit} or \textit{somatic phraseology} for the vocabulary; however, \textit{somatic reference} and \textit{somatic expression} can be used more generally to express even non-language signals associated with the body.}} 
%While BPMs are founded on a relatively simple connection 
Even though BPMs are a relatively simple method for detecting embodiment-related language, we propose them as a useful initial tool for investigating the relationship between language and our embodied selves in everyday language. 
% relatively simple compared to more sophisticated traditional identifiers of embodiment such as BLAH,
% between language and embodiment---words with a semantic association to the human body---
As such, we use \sw{BPMs in this paper for an exploratory investigation of how words which are semantically related to the human body can provide interesting insights in natural language use.} % through existing corpora. 

% In this work, 
We introduce two novel BPM corpora, Spinn3r$_{\rm \it BPM}$ (a corpus of blog posts) and TUSC$_{BPM}$ (a corpus of tweets), and conduct a variety of experiments to show how they can be used to address three areas of inquiry on the connection between embodiment and natural language. First, we investigate the prevalence of body-part related words in everyday language, as well as how frequency of these words differs across different factors such as medium, gendered pronoun usage, time, and place.
% To do this, we look at the prevalence of BPMs in our corpora, the most frequently referred-to BPMs, possessive pronouns most likely to precede BPMs, 
% and how different BPM categories differ in terms of the context that surrounds them. 
Second, we look at the relationship between BPMs and affect, motivated by the growing pool of research which posits that emotions originate from interpretations of physiological signals. Lastly, we propose and test the hypothesis that the %extent 
\sm{degree of prevalence} of BPMs in social media is indicative of aggregate-level health outcomes. 

% \sw{Our primary aim is to conduct the first large-scale analysis of Body Part Mentions (BPMs) in natural language. Rather than isolating specific mechanisms, our goal is to probe whether BPMs carry diverse, meaningful associations to sociological factors and emotional associations in natural language which could warrant further study.}
% Commenting to save space :: That is, markedly more frequent mentions of body parts in social media posts in one region compared to another could be indicative of different health outcomes.
% evaluate the relationship between regional BPM use and health outcomes using available metadata in our corpus.

A better understanding of embodiment through language could support \sm{work in} a range of research areas, especially NLP in health domains (where BPM-heavy corpora are frequent 
%and their interpretations carry  consequences for patients 
\cite{chaturvedi2023development}). the growing wave of interest in integrating embodiment within computational models of language (especially text-based LLMs, which rely on BPMs to access information about human embodiment, and struggle with many benchmarks of human cognition due to their lack of embodiment \cite{chemero2023llms}).

% , and cross-linguistic and cross-cultural studies on embodiment (which are often limited by difficulties in accessing a wide range of participants from different backgrounds) \cite{koczy2023embodiment}.

\sm{All of our code and data is available at our project repository.\footnote{\url{https://www.github.com/sohpei/bpms/}}}

\section{Related Work}

\subsection{Embodiment, Affect, and Health}
\sm{According to} the \textit{theory of constructed emotion}
%, (introduced by Lisa Feldman Barrett 
\cite{barrett2017theory},
%, a leading scholar in the affective sciences), 
% proposes that 
emotions emerge as interpretations of our bodies' physiological signals. A key \sm{supporting} argument for this theory is the wide range of research establishing a strong connection between \textit{interoception}---the ability to feel internal bodily sensations \cite{craig2002you}---and emotional wellbeing. Better interoceptive awareness has been shown to positively correlate with better emotional regulation \cite{zamariola2019relationship}, emotional decision-making \cite{dunn2010listening}, and emotional granularity (the ability to distinguish different emotions) \cite{feldman2024neurobiology}. These results indicate that awareness of our bodily experiences is crucial to our emotional welfare %.
and dysfunctional interoception is %possibly a major 
a contributor to a variety of mental health conditions \cite{khalsa2018interoception}.

Embodied experiences %also 
manifest frequently in everyday language. For example, descriptions of bodily experiences are frequently found in storytelling contexts \cite{gallese2011stories}. But even outside of their role as explicit %ly 
physical referents, % body-related words 
BPMs may reveal a deeper connection to embodied phenomena. The  \textit{theory of conceptual metaphor} suggests that metaphors are a fundamental cognitive process \cite{lakoff2008metaphors}. According to this framework, metaphors help us understand abstract concepts by mapping them onto concrete, physical experiences. Treated this way, the widespread usage of body part words in language---even those applied in abstract contexts---are important for understanding how natural language is connected to the physical world from which it originates. Scholars in affective theory have argued that this is why \sm{expressions of} intensely emotional experiences  often use body parts and actions as a metaphor (e.g., \textit{my heart is broken}, \textit{weight lifted off my shoulders}); they are effective ways of grounding subjective experience in a shared embodied reality \cite{kovecses2003metaphor}.

% Studies have confirmed that metaphorical language is associated with more emotion than non-metaphorical language \cite{mohammad2016metaphor}.

% ^thought this would be cool to mention but I'm still struggling to find the link:")

% Embodied experiences manifest frequently in language. Bodily experiences form an important component of our written narratives \cite{gallese2011stories}, individuals often use spatially orientated language to conceptualize abstract ideas \cite{lakoff2008metaphors}, and cross-linguistic studies have revealed how sensory experiences can shape linguistic expressions [TK cite]. [TK cite - embodied interaction: language and body in the material world].

\subsection{Embodiment and NLP}

Language plays a key role in understanding the relationship between embodiment and affect. Emotional granularity is measured from the ability to identify and distinguish different emotions using words \cite{tan2022emotional}, and interoception is often measured through an individual's ability to describe their internal state \cite{desmedt2022measures}. 
%Notably, a 
A range of NLP projects have taken an interest in body part words for specific applications, such as identifying gendered representations in literature \cite{silva2024words}, mapping bodily sensations for healthcare applications \cite{wang2019mapping}, or building computational methods for detecting body parts involved in emotional processes to improve machine emotion recognition \cite{zhuang-etal-2024-heart}. \sw{We note that other related work in NLP largely uses body-related language to answer adjacent research questions in specialized settings, rather than investigate the significance of BPMs themselves. Since this is the first-ever work investigating the significance of BPMs in \textit{everyday language} using NLP methods, we aim to investigate whether the general relationship between affect and body parts suggested and observed in laboratory environments (i.e., the relationship between affect and body part words suggested by the theory of conceptual metaphor, or the relationship between affect and described bodily experiences suggested by the theory of constructed emotion), can be corroborated using textual corpora}.

% Many data-driven tools have already been developed to investigate emotions in natural text at scale, which provide a rich opportunity to see how these phenomena can be investigated in everyday online language. 

% \sm{[ADD some sentences about existing emotion lexicons.]}
\sm{Many word--emotion association lexicons have been created: mostly for English: e.g., \newcite{bradley1999affective},  the NRC Emotion Lexicon \cite{Mohammad13,mohammad-turney-2010-emotions},
\newcite{warriner2013norms}, The NRC VAD lexicon \cite{mohammad-2018-obtaining}; but also for some other languages: e.g., \newcite{moors2013norms} for Dutch,  \newcite{vo2009berlin} for German, and \newcite{redondo2007spanish} for Spanish.
The NRC Emotion Lexicon includes entries for whether a words is associated with eight categorical emotions: joy, sadness, fear, anticipation, anger, trust, disgust, and surprise (the \newcite{Plutchik80} set).
It includes entries for
about 14,000 words.\footnote{http://saifmohammad.com/WebPages/NRC-Emotion-Lexicon.htm}
The NRC VAD lexicon \cite{mohammad-2018-obtaining} has valence, arousal, and dominance associations for about 20,000 English words.\footnote{http://saifmohammad.com/WebPages/nrc-vad.html}
(Version 2 of the NRC VAD Lexicon was released recently, and it includes entries for about 44k words and 10k multiword expressions.)}

In this work, %paper, 
we primarily use lexicons %for 
of word--emotion associations---which can be used to create accurate emotion arcs in streams of text \cite{teodorescu2023evaluating} or
effectively distinguish emotional granularity \cite{vishnubhotla2024emotion} 
---as a computationally inexpensive and interpretable method for beginning to look at the relationship between affect, language, and embodiment.

% \sw{We note that this is the first-ever work investigating the significance of BPMs in everyday language using NLP methods, so our related work largely references NLP papers that use body-related language to answer adjacent research questions rather than the significance of BPMs themselves.  We aim, in the rest of the paper, to show that this significance exists through the diverse and interesting relationships which exist between BPM usage and various meaningful factors.}

\section{BPM Corpora}

% the \textit{Blog BPM corpus} and two \textit{Tweet BPM Corpora}. All three corpora were created from existing personal blog and tweet corpora by extracting sentences that included mentions of body parts. The Blog BPM corpus was compiled from the Spinn3r ICWSM 2011 dataset \cite{burton2009spinn3r, burton2011spinn3r}. 
% The Tweet BPM Corpus was compiled by selecting tweets from TUSC (a dataset of geo-located tweets from US and Canada from 2015 to 2021) \cite{vishnubhotla-mohammad-2022-tusc}. Therefore we will also refer to the two corpora as Spinn3r$_{\rm \it BPM}$ and TUSC$_{BPM}$, respectively.
%
%\subsection{Filtering for BPMs}
%self-re
For this work, we investigate two mediums for online, everyday language: blog posts and tweets. 
% We define an \textit{instance} from a corpus asividual sentence from a blog post or an individual tweet. 
We consider each sentence from a blog post and each individual tweet from the tweet corpora as \sm{an} instance. %\sw{, which aligns with prior methodologies for segmenting text for emotion detection \cite{das2009sentence} \cite{gaind2019emotion}}
% Instances that included tokens from the BPM list were then extracted from the 
We created BPM corpora by extracting all instances that included at least one word referring to a body part from the
Spinn3r personal blog datasets \cite{burton2009spinn3r, burton2011spinn3r}, and the two TUSC datasets:
% of geo-located tweets posted between 2015 and 2021, 
TUSC$_{\rm \it ctry}$ (where tweets are geo-located to either the United States of America or Canada) and TUSC$_{\rm \it city}$ (where tweets are geo-located to cities in North America) \cite{vishnubhotla-mohammad-2022-tusc}. 
This results in three final BPM corpora: Spinn3r$_{\rm \it BPM}$, TUSC$_{\rm \it ctry-BPM}$, and TUSC$_{\rm \it city-BPM}$. 
We will refer to the corpora made up of the rest of the instances as 
Spinn3r$_{\rm \it noBPM}$ and TUSC$_{noBPM}$. 

% We use the same body parts word lists as in 
\sm{We compile a list of body part words by including all the terms in the list used by}
\citet{zhuang-etal-2024-heart}, which extracts BPM samples from the Spinn3r corpus to annotate samples for the presence of explicitly embodied emotion.\footnote{%\url{
https://www.collinsdictionary.com/us/word-lists/body-parts-of-the-body}%}
\footnote{%\url{
https: //www.enchantedlearning.com/wordlist/body.shtml}%}
Additionally, we include plural forms of 
terms (e.g., \textit{hearts, hands, eyes,} etc.).
%in our final BPM list. 
We refer to the body part word forms %included in our list 
as \textit{BP word types}. 
The list of 295 BP word types we used is 
available on the project webpage.
%included as a part of our released dataset. 

%There are t
Two issues need to be addressed when working with BPM corpora to study embodiment. First, some BP word types are ambiguous (some BPM instances 
may not actually be referring to a %human 
body part: e.g, `I will be \textit{back}'). Second, it is useful to distinguish between the speaker referring to their own body vs.\@ the speaker referring to someone else's body. We are especially interested in mentions of one's own body as a possibly useful indicator for %interoception and 
well-being. 
While we are interested in the insights that all BPMs, including those with a more abstract/metaphorical connections to the human body, can offer on the relationship between embodiment and language, it would also be useful to distinguish instances where the speaker refers directly to their own body parts. and so we use a simple solution that effectively addresses both of the issues raised above. 
We created three subsets for each BPM corpus that only include the BPM instances preceded by possessive pronouns `my', `your', and `his/her/their', respectively, and call these instances: \textit{possessed BPMs}
% \footnote{We exclude the pronoun "our" because it is often used to refer to general body part behavior, but keep the pronoun "their" since this can be used to refer to a person's body part in a gender-neutral way, and plural references to a group of people's body parts usually specify a specific group's body parts. We also exclude the pronoun "its" because it often refers to non-human entities.}
This helps us create separate corpora for first-, second-, and third-person references to body parts, and also excludes a vast majority of mentions that are not explicit references to body parts. For example, `I will be \textit{back}' would not be included in any of these corpora, but `my \textit{back} hurts' will be. We find that this approach delivers a high number of references to an individual's body part (92\% of 100 manually inspected instances). While this approach sacrifices recall by excluding possible references to body parts without possessive pronouns, we benefit from higher precision.\footnote{For our experiments, we do not need all BPM instances, but rather just a large sample.} Additionally, we also conduct some experiments with all BPM instances (the higher-recall and lower-precision corpus).
% the resulting dataset is much more suitable to the analysis of BPMs and affect since these BPMs can be clearly attributed to a particular body which possesses the BPM.

% For this work we define an instance as an individual sentence from the blog posts in Spinn3r and as an individual tweet from TUSC. 
% containing at least one BPM in 
% Spinn3r$_{\rm \it BPM}$, TUSC$_{\rm \it ctry-BPM}$, and TUSC$_{\rm \it city-BPM}$ and use all other instances in Spinn3r$_{\rm \it noBPM}$ and TUSC$_{noBPM}$. 

% This results in $8,371/80,379$ sentences being included in Spinn3r$_{\rm \it BPM}$, $6,710,660/104,575,991$ sentences being included in TUSC$_{\rm \it city-BPM}$, and $231,577/3,181,879$ sentences being included in TUSC$_{\rm \it ctry}$. 

\subsection{Emotion-Annotated BPM Corpus}
%Building on this foundation, 
We also leverage a specialized subset of Spinn3r, previously released by \newcite{zhuang-etal-2024-heart}. This subset  contains sentences with BPMs annotated for the presence of  explicitly embodied emotion, which is defined as "the physical experience of an emotion via our body" (i.e., \textit{"Julie pouted and rolled her eyes"} is annotated as containing embodied emotion, but \textit{"Frank breathed heavily through his mouth after his run"} is not). We refer to this subset in the rest of this paper as Spinn3r$_{\rm \it BPM-Zhuang}$. We extend this work by creating the first human-annotated dataset that explicitly identifies BPM ownership (whether the BPM refers to the speaker's body or not) and the emotion of the BPM owner (joy, fear, etc.) as inferred by a human reader.
% Through this dataset, we forward a novel method of emotion annotation that allows for standardized inference of emotions beyond the speaker, which better reflects the diversity of relationships from which individuals detect sentiment in the real world. 

%Both of these datasets are released for public usage.
For full details of all datasets and data collection methods see \Cref{sec:appendix_datasets}. Further details for the emotion annotation process, which includes a set of quality checks and aggregation methods for final emotion scores, can be found in \Cref{sec:annotation}.

% \subsection{Tweet BPM Corpus}

% We extract BPMs from TUSC, a dataset of 45 million geo-located tweets posted between 2015 and 2021 from the US and Canada \cite{vishnubhotla-mohammad-2022-tusc}. 
% We filter our corpus for tweets that contain at least one BPM token and discard BPMs that have fewer than 0. 1\% instances in relation to all instances with a BPM token. We find 6,937,472 out of 107,757,870 instances that have at least one BPM token (226,812 out of 3,181,879 from the \textit{ctry} (country) part and 6,710,660 out of 104,575,991 instances from the \textit{city} part).

% \subsection{Blog BPM Corpus}

% We extract BPMs from Spinn3r, a blog post dataset a dataset of 

% blog posts, tweets
% Spinn3r, TUSC (TUSC-city, TUSC-ctry)
% Spinn3r-BPM, TUSC-BPM
% Spinn3r-BPM-SpEmo, TUSC-BPM-SpEmo

% \section{Emotion Annotations of BPM Corpora}

% Existing emotion datasets may not have many BPMs therefore not doing this.

\section{Research Questions About BPMs}

Despite substantial evidence from medical research and psychology that points to connections between the mind and the rest of the body, as well as the connection between interoception and emotional well-being, there is little quantitative work using language to explore this connection. In this section and the next, we make use of large amounts of social media data, massive word--emotion lexicons, and the emotion-annotated corpus described in the previous section, to examine questions on how, when, and in what context we refer to our body parts in text (this section) and whether different body parts mentions tend to be used in different emotional contexts (next section). Since the questions in this section are relevant to the \textbf{B}ody, we will index them as B1, B2, etc. The questions in the next section are related to \textbf{B}ody and \textbf{A}ffect, so we will index them as BA1, BA2, etc.\\[3pt] 
\noindent \textbf{B1. To what extent do we use body-related words? While it may be difficult to get natural conversational data for privacy reasons, how often do we mention body parts in social media?}\\
\noindent \textit{Method:}  We did not find any past research on how common BPMs are in language. We do not even have a sense of the magnitude: do they occur in 0.01\%, 0.0001\%, 1\%, 10\%, \sm{60\% of utterances, or something else}?
To examine the extent to which body-related vocabulary is used in online language, we calculated the number of instances containing at least one BPM.\\ 
\noindent \textit{Results:} See \Cref{tab:BPM_instances}. We find that a substantial proportion of instances contain at least one BPM: 
% 10.41\% of blog post sentences in the Spinn3r dataset, 6.42\% and 7.28\% 
10.4\% of blog post sentences in the Spinn3r dataset, 6.4\% and 7.3\% 
of tweets in in TUSC$_{\rm \it ctry}$ and TUSC$_{\rm \it city}$, respectively.\\
\noindent \textit{Discussion:} The consistently high proportion of BPM-containing instances across our corpora emphasizes the ubiquity of body part references in online English text.
The markedly larger proportion \sm{of} samples containing BPMs in blog sentences than tweets implies that usage of 
BPMs may be more pervasive in personal narratives and longer-form text than in tweets.\\[3pt]
\begin{table}[t]
\centering
\resizebox{\columnwidth}{!}{
    \begin{tabular}{lrrr}
     \specialrule{.12em}{.05em}{.08em}
     \textbf{Corpus}    & \textbf{S}  & \textbf{T}$_{city}$ & \textbf{T}$_{ctry}$\\ 
     \textbf{Instances}      & \small{\textit{(80,379)}}  & \small{\textit{(104,575,991)}} & \small{\textit{(3,181,879)}}\\ 
      \textit{<BPM>} instances  & 10.4 & 6.4 & 7.3 \\ 
     \specialrule{.12em}{.05em}{.08em}
    \end{tabular}
}
\caption{B1 - Percentage of instances in each corpus with at least one BPM.
 S = Spinn3r. T = TUSC.}
\label{tab:BPM_instances}
\end{table}
\noindent \textbf{B2. To what extent do we talk about our own body parts (i.e., my BPM) versus others'  body parts (i.e., your/his/her/their BPM)?}
\begin{table}[t]
\centering
%\small{
 \resizebox{0.9\columnwidth}{!}{
    \begin{tabular}{lrrr}
     \specialrule{.12em}{.05em}{.08em}
     \textbf{Corpus}    & \textbf{S}$_{bpm}$  & \textbf{T}$_{city\_bpm}$ & \textbf{T}$_{ctry\_bpm}$\\ 
     \textbf{Instances}       & \small{\textit{(8,371)}}  & \small{\textit{(6,710,660)}} & \small{\textit{(231,577) }}\\ 
      Possessed BPM & 28.9  & 31.9 & 26.3 \\
      \hspace{0.5cm} \textit{``my <BPM>''} & 16.6  & 19.2 & 15.8\\
      \hspace{0.5cm} \textit{your <BPM>} & 6.5 & 6.6 &  5.5\\
      \hspace{0.5cm} \textit{his <BPM>} & 2.5  & 3.3 & 2.7 \\
      \hspace{0.5cm} \textit{her <BPM>} & 2.2  & 1.6 & 1.4 \\      
      \hspace{0.5cm} \textit{their <BPM>} & 1.2  & 1.3 & 1.0 \\
     \specialrule{.12em}{.05em}{.08em}
    \end{tabular}
}
\caption{B2 - Percentage of instances containing a possessed BPM out of overall BPM samples.}
% in each dataset.}
\label{tab:BA2.possessedBPM}
\vspace*{-3mm}
\end{table}

\noindent \textit{Method:} Although BPMs can be used to study the general significance of body part words in language, we are also interested in examining the extent to which these body parts can be attributed to a particular person (e.g., \textit{her heart}) vs.\@ body part words that %cannot be 
are not attributed to a human possessor (e.g., \textit{the heart of the matter}). To do this, we introduce the concept of \textit{possessed BPMs}, which can be attributed to a particular body using a possessive pronoun. We look at three general categories of possessed BPMs: first person instances including \textit{``my <BPM>''}, second person instances including \textit{``your <BPM>''}, and third person instances including \textit{``his/her/their <BPM>''}. In each corpus, we determine the frequency of instances containing at least one instance of a possessed BPM.\\
\noindent \textit{Results:}  \Cref{tab:BA2.possessedBPM} shows the results. In each corpus, the most common possessed BPM is \textit{``my <BPM>''}, and there are more instances containing \textit{``his <BPM>''} than \textit{``her <BPM''}.\\
\noindent \textit{Discussion:} By evaluating different distributions of  possessed BPM types, we can begin building an understanding of \textit{whose} body parts are most often referred to in conversation. 
Our results indicate that individuals are more likely to discuss their own bodily experiences in online discourse than that of others. The higher frequencies of  \textit{``his <BPM>"} over \textit{``her <BPM>"} instances are also interesting, considering well-established theories that women's bodies are more heavily discussed and scrutinized in popular media \cite{bordo2023unbearable}---our results indicate that in spite of this, men's bodies may be referred to more often in everyday speech.\\[3pt]
%% The relatively high frequency of yourBPM in the blog datasets, and markedly lower number in tweets, also implies that BPMs are more likely to be used in advice-giving and distancing contexts in longer-form text than in tweets, as according to previous work on second-person pronoun use \cite{pennebaker2011secret}.
\noindent \textbf{B3. Which of our body parts do we refer to most often? Do we refer to our body differently in different online contexts?}

\noindent \textit{Method:} To answer this, we calculate the frequencies of \textbf{each individual} BP word type preceded by the possessive pronoun "my" (i.e., \textit{my <BPM>}).

\noindent \textit{Results:}  
% \noindent \textit{Discussion:} 
 We find that there are certain BP word types that appear very frequently in all corpora---with twelve \textit{``my <BPM>''} word types being shared in the top twenty across all corpora. However, we also observe variation in frequencies across corpora, indicating that we describe our body in different ways across different online contexts.
Across all corpora, \textit{my heart} and \textit{my head} are among the most frequently mentioned \textit{``my <BPM>''} word types. These body parts are likely central sources for people's basic understandings of their embodied experiences, which is reflected in the prevalence of common figurative expressions such as \textit{my heart is broken} and \textit{my head hurts}. We also find that the blog dataset has a much stronger representation of body parts that are strongly related with human senses, such as \textit{my eyes} (10.23\% vs.\@ 1.40\%), \textit{my ears} (1.19\% vs.\@ <0.1\%), and \textit{my hands} (3.32\% <0.1\%). This suggests that the personal narratives in blogs may be more focused on sensory, everyday experiences. Additionally, \textit{my hair} and \textit{my face} appear much more frequently in the TUSC tweet datasets than in the Spinn3r blog datasets, likely a result of personal grooming and appearance being more prevalent in %short-form 
social media updates. This rich divergence between common \textit{``my <BPM>''} word types implies that users refer to their body differently when expressing themselves in different online mediums.
(\Cref{tab:BPM_percentages_overall} in the Appendix shows the top 20 \textit{``my <BPM>''} types in each corpus.)\\[3pt] 
% with their frequencies.)
\begin{figure}[t]
    \centering
    \includegraphics[width=\columnwidth]% ,height=2.4cm]
    {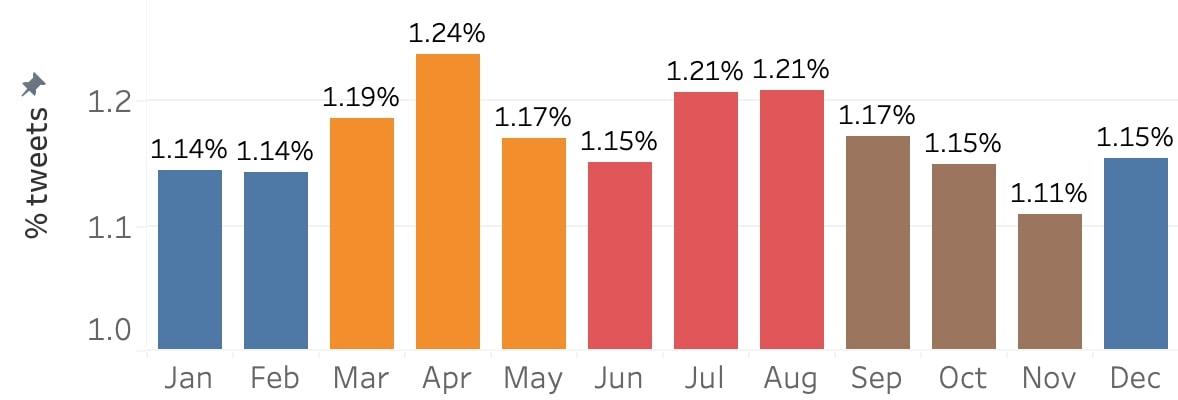}
    \vspace*{-4mm}
    \caption{B4 - TUSC$_{\rm \it ctry}$ - \% of tweets with at least one \textit{``my <BPM>''} by month. Colored by season in USA.} 
    % \footnote{\url{https://en.wikipedia.org/wiki/Season}} (winter: dark blue, spring: orange, summer: red, fall: brown).}
    \vspace*{-1mm}
    \label{fig:BPMs_by_month_tusc_country}
\end{figure}
\begin{figure}[t]
    \centering
    \includegraphics[width=0.85\columnwidth] %,height=2.4cm] 
    {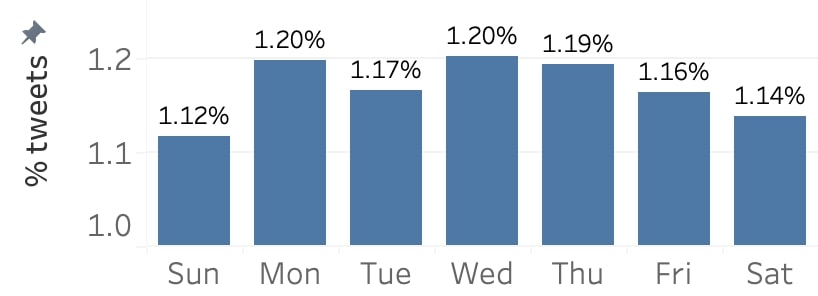}
    \vspace*{-1mm}
    \caption{B4 - TUSC$_{\rm \it ctry}$ - \% of tweets with at least one \textit{``my <BPM>''} for different weekdays.}
    \vspace*{-3mm}
    \label{fig:BPMs_by_weekday}
\end{figure}
\begin{figure*}[t]
    \centering \includegraphics[width=0.9\textwidth]{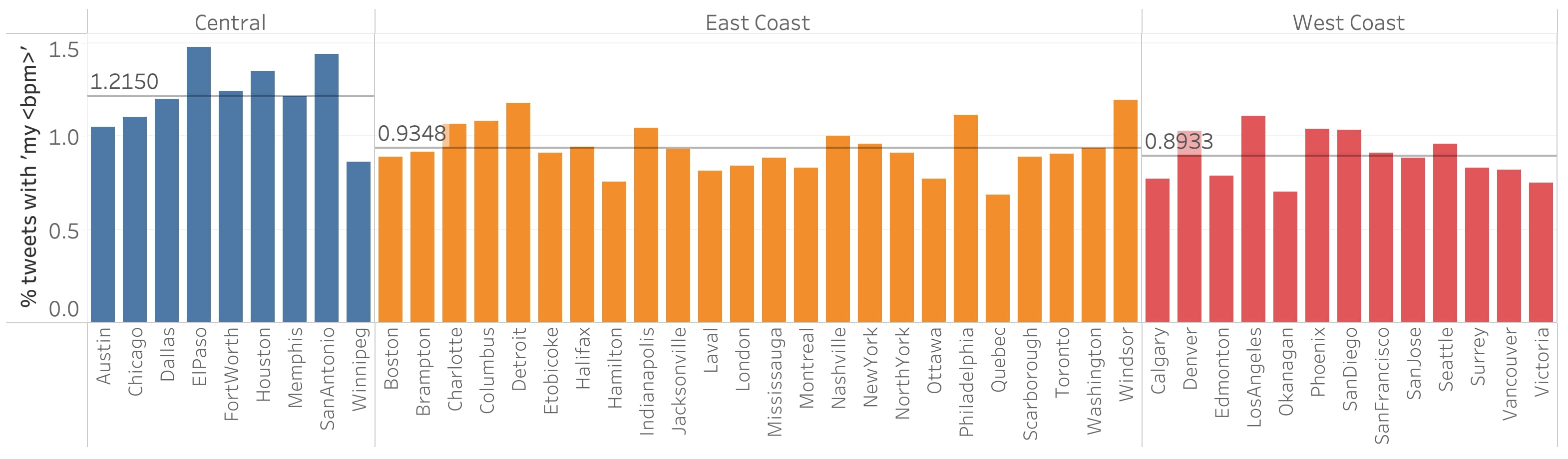}
    \vspace*{-1mm}
    \caption{B5 - TUSC$_{\rm \it city}$ - \% of tweets with at least one ``my <BPM>'' for different cities.}
    % in the east coast, west coast, and central region of Canada and USA in 2020-2021.}
    \label{fig:BPMs_by_city}
    \vspace*{-3mm}
\end{figure*}

\begin{figure}[t]
    \centering
    \includegraphics[width=0.47\textwidth]{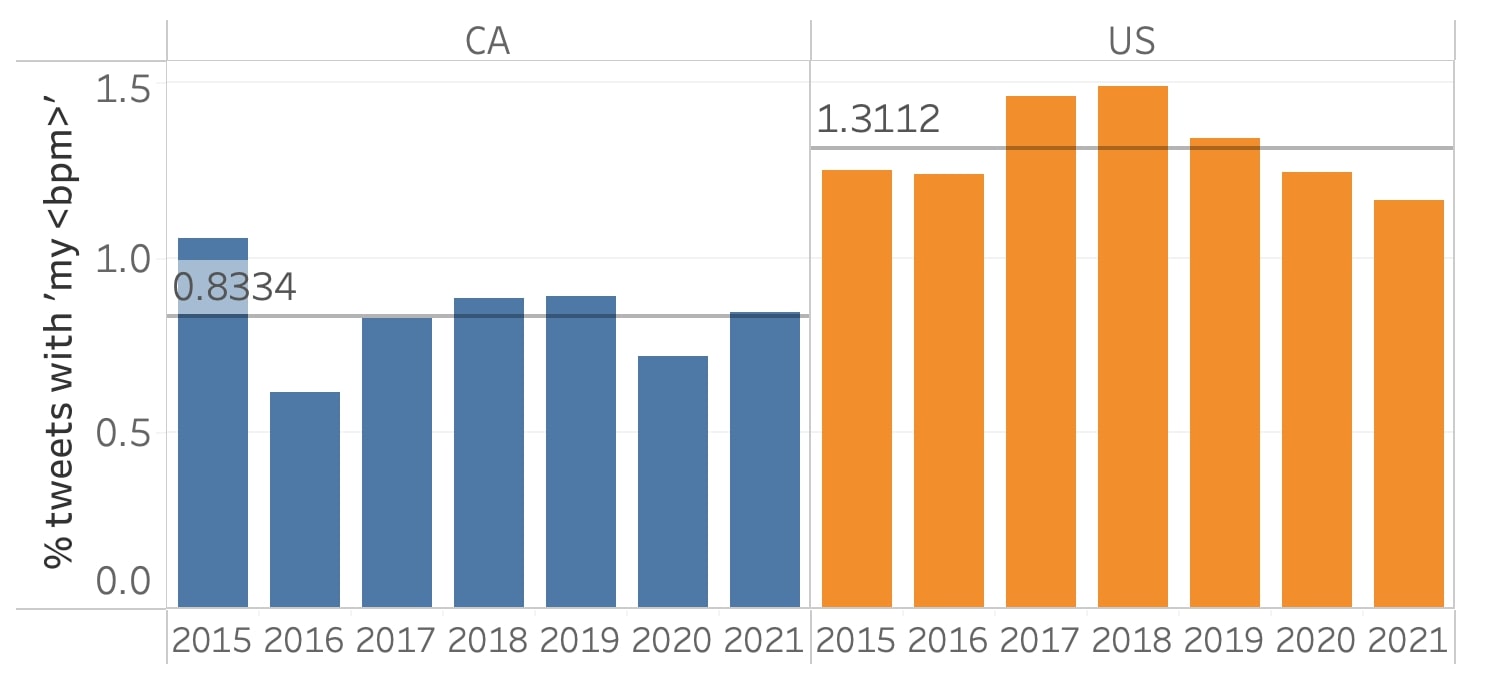}
    \caption{B5 - TUSC$_{\rm \it ctry}$ - \% of tweets with at least one ``my <BPM>'' for Canada and USA from 2015 to 2021.}
    \label{fig:BPMs_by_country_by_year}
\end{figure}

\noindent \textbf{B4. Does the time of the week/year impact %whether 
\sm{the extent to which} we refer to our body?}

\noindent \textit{Method:} Each sample in the TUSC$_{\rm \it ctry}$ dataset has %the exact timestamp 
\sm{a timestamp indicating the exact time}
at which it was posted. We use this data to examine whether \textit{``my <BPM>''} usage is higher or lower at different times.\\ %cales: day of the week, month, season (by mapping the months to seasons that are experienced in the northern hemisphere), and year.\\
\noindent \textit{Results:} Frequency of instances containing \textit{``my <BPM>''} by day of week and month are shown in Figures \ref{fig:BPMs_by_month_tusc_country} and \ref{fig:BPMs_by_weekday}.
We find that \textit{``my <BPM>''} instances peak during the summer and spring, decline steadily during the fall, and then stay relatively low during the winter. We also find that \textit{"<BPM>"} instance frequency is highly dependent on different days of the week, rising steadily from Sunday to Wednesday and then declining from Wednesday to Saturday. We find a statistically significant decline in BPM usage from April onwards in the year and from Monday onwards throughout the week. More details on statistical tests can be found in \Cref{sec:time_statisticaltests}. \\
\noindent \textit{Discussion:} The seasonal differences in \textit{``my <BPM>''} usage in warmer months indicates that factors such as temperature, sunlight, and time spent outside could affect awareness and expression of %an individual's 
one's bodily experiences. The weekly rise of referral to one's own body parts may reflect a renewed engagement with structured activities as the work week begins, while the decline could indicate fatigue or decreased energy as the week progresses, making it difficult to cultivate bodily awareness, consistent with documented patterns of weekly fatigue cycles in organizational research \cite{zijlstra2008weekly}. These results indicate that embodied language use is not static, but responds to environmental and social rhythms. \\ [3pt]
%This could warrant future research on the effect of temperature, routine, and other environmental factors on interoception and bodily awareness.\\[3pt]

\noindent \textbf{B5. Do individuals in different regions refer to their bodies at different frequencies?}\\
\noindent \textit{Method:} We take advantage of the geotagged meta data available for the TUSC tweets %in TUSC$_{\rm \it city}$ and TUSC$_{\rm \it ctry}$ 
to evaluate the regional proportion of \textit{``my <BPM>''} tweets.\\
\noindent \textit{Results:} Figure \ref{fig:BPMs_by_city} shows BPM use by city. We find that \textit{``my <BPM>''} instances are used more in central cities than coastal cities. We also find that \textit{``my <BPM>''} instances are more frequent in American tweets than Canadian tweets in the TUSC$_{\rm \it ctry}$ dataset, as shown in \Cref{fig:BPMs_by_country_by_year} (TUSC$_{\rm \it city}$ dataset). \sw{We find also find that the city of a post has a statistically significant effect on the usage of BPMs in a city. We also find that country has a slightly smaller but still statistically significant effect on usage, but that this pattern also persists across the different months represented in our dataset. Details on statistical tests, as well as a figure showing that usage difference between USA and Canada is consistent across months, can be found in \Cref{sec:bpm_ctry}.}\\%%\Cref{fig:BPMs_by_country_by_month} in the Appendix displays the frequency of \textit{``my <BPM>''} tweets across US and Canada for each month from 2015 to 2021.\\
\noindent \textit{Discussion:} These findings suggest that regional differences influence how individuals refer to their bodies, potentially reflecting broader cultural, social, or environmental factors. Future research could explore how variables such as climate, healthcare access, or local discourse shape how individuals discuss their body in different regions.

% For the TUSC$_{\rm \it ctry-BPM}$ dataset, there are:
% \begin{itemize}
%     \item 56 BPM types for ``my'' pronoun with $>0.1\%$ occurrence
%     \item 54 BPM types for ``your'' pronoun with $>0.1\%$ occurrence
%     \item 108 BPM types for ``his/her/their'' pronoun with $>0.1\%$ occurrence
% \end{itemize}
% For the Spinn3r$_{\rm \it BPM}$ dataset, there are:
% \begin{itemize}
%     \item 57 BPM types for ``my'' pronoun with $>0.1\%$ occurrence
%     \item 40 BPM types for ``your'' pronoun with $>0.1\%$ occurrence
%     \item 131 BPM types for ``his/her/their'' pronoun with $>0.1\%$ occurrence
% \end{itemize}
\section{Research Questions on BPMs--Affect}
% \begin{table}[t]
%     \centering
%     \begin{tabular}{lrrr}
%         \toprule
%         & \textbf{Spinn3r$_{\rm \it BPM}$}  & \textbf{TUSC$_{\rm \it ctry}$} & \textbf{TUSC$_{\rm \it city}$}\\ \midrule
%          my <BPM> & 1.45 & 1.14 & 1.01 \\
%          your <BPM> & 1.11 & 0.41 & 0.35 \\
%          his <BPM> & 0.61 & 0.20 & \\
%          her <BPM> & 0.19 & 0.10 & \\
%          their <BPM> & 0.29 & 0.07 & \\
%          \bottomrule
%     \end{tabular}
%     \caption{B1. Frequency of BPMs preceded by first-person, second-person, and third-person possessive pronouns.}
%     \label{tab:possessive_pronouns}
% \end{table}

% These research questions concern the relationship between affect and body part mentions (BPMs). 
% In this section, we take a special interest in 
The primary goal of our work is to explore how language can shed light on the connection between the body, emotion, and well-being. In this section we explore how BPMs are associated with emotions. 
% possessed BPMs, especially 
% We focus on possessed BPMs, and especially myBPMs (\textit{my arm, my head,} etc.) 
% because such instances are highly likely to refer to one's own body parts.
% since these instances guarantee a speaker referring to their own embodied selves. 
We explore this question using % the BPM corpora, 
% (including the emotion annotated corpus discussed in Section 3)  
emotions associated with words that co-occur with BPMs (using large word--emotion association lexicons)
as well as perceived emotions of the speaker (using the new human-annotated Spinn3r$_{\rm \it BPM-Zhuang}$ dataset we introduced earlier). \sw{In this section, we take a special interest in samples including \textit{"my <BPM>"}, since our emphasis is rooted in theories of embodied emotion and health psychology, which suggest that references to one's own body are more likely to reflect internal emotional and physical states.}%\\[3pt]
%We are interested in two approaches for uncovering associated affect with BPMs. Firstly, we are interested in the \textit{emotional signature} of body part words, where we define as the co-occurence of BPMs with emotion-associated words. Secondly, we look at the \textit{emotions of the BPM possessor}, where we use the subset of Spinn3r which we have annotated for emotions expressed by owners of particular BPMs. \\[3pt]
% \noindent \textbf{BA1. Does body part-related language produce a different emotional signature than text with no body part mentions?}

\noindent \textbf{BA1. Do posts with body part mentions have markedly different emotional associations?}\\[3pt]
\noindent \textit{Method:} This question aims to shed light on whether the relationship between emotion and embodiment manifests in social media text. 
% \sw{Our research question focuses on identifying general patterns of emotional language associated with Body Part Mentions (BPMs)—which are themselves extracted using a lexicon-based method—across large-scale corpora. To address this, we deliberately use lexicon-based methods, which—despite known limitations in handling context, are a well-established and computationally inexpensive method which has been shown to effectively capture aggregate emotion arcs (changes in emotions) across large samples \cite{teodorescu2023evaluating} as well as effectively capture emotional well-being phenomena on large corpora \cite{vishnubhotla2024emotion}. We chose not to use more complex emotion detection models (e.g., ML-based classifiers) in order to maintain alignment with our broader, descriptive research goal. While such models could provide more fine-grained or context-sensitive emotion detection, they are often less interpretable and highly corpus-dependent.}
In this experiment, we look at the proportion of samples (tweets/blog posts) containing at least one word associated with various emotion categories: \textit{anger, anticipation, disgust, fear, joy, sadness, surprise, and trust} from Plutchik's set of emotions \cite{plutchik2001nature},
and high or low \textit{valence (positive--negative), arousal (calm--sluggish), and dominance (in control--out of control)}.
We obtain the word--emotion associations from the NRC Emotion Lexicon \cite{Mohammad13,mohammad-turney-2010-emotions} and the NRC VAD Lexicon \cite{mohammad2018obtaining}. 
We compute these proportions \sw{for various BPM categories:}  \textit{``my <BPM>''}, \textit{"his/her/their <BPM>"}, \textit{"your <BPM>"}, as well as the 
no BPM corpora. 
% \sw{Although this filtering  may not capture all BPMs attributable to a human body in a text, our primary goal in using possessive pronouns was to ensure that the emotional signal associated with BPMs was interpretable and attributable to a subject, which allows us to investigate relationships between bodily referral, language use, and affect, as suggested by the theory of constructed emotion.}
\begin{figure}
    \centering
    \includegraphics[width=\columnwidth]{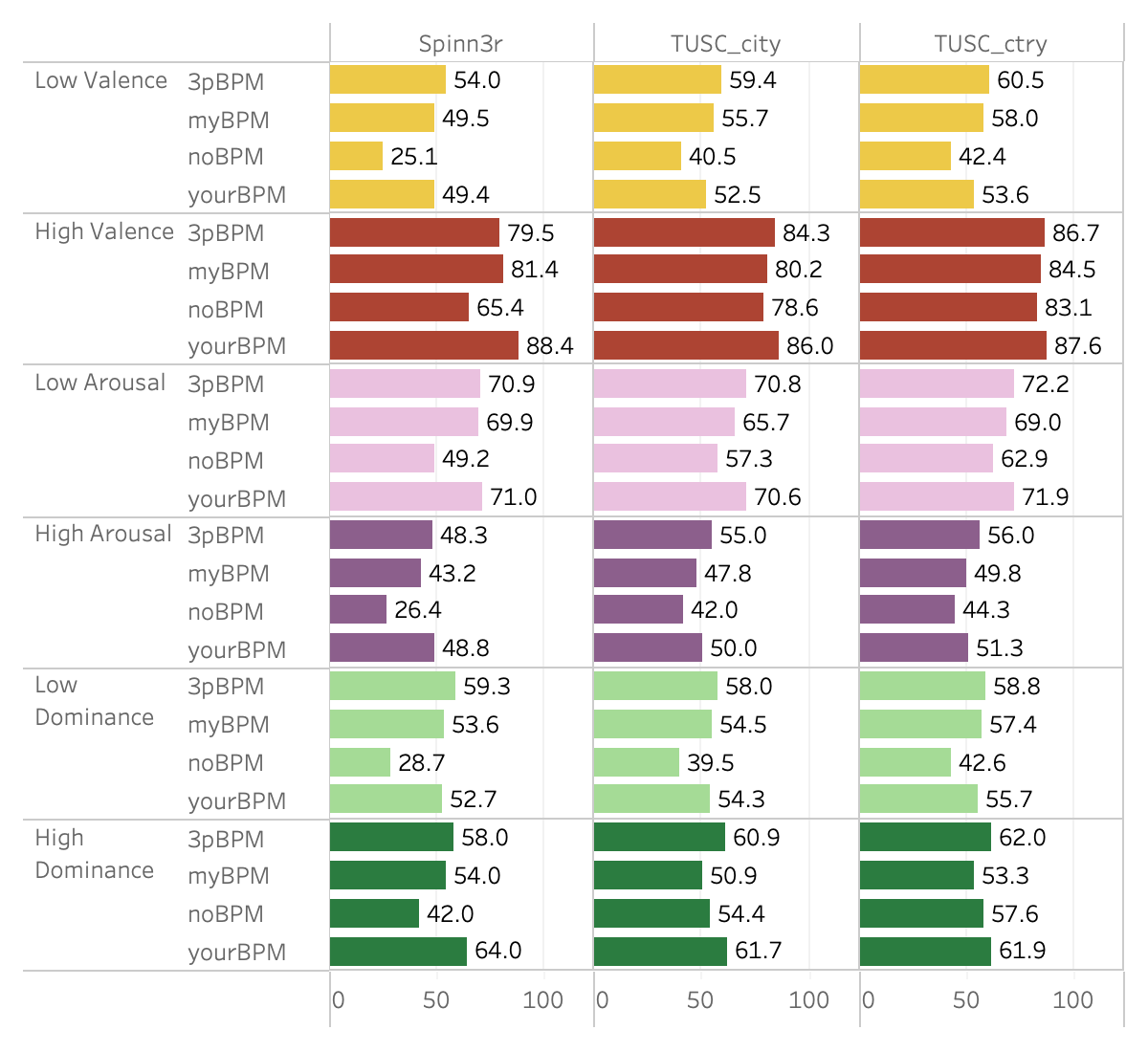}
    \caption{BA1 - Percentage of sentences with at least one high or low valence, arousal, or dominance word (according to the NRC VAD lexicon) in each corpus in myBPM, yourBPM, 3pBPM, and noBPM categories.}
    \label{fig:VAD_pronouns-fig.png}
    \vspace*{-3mm}
\end{figure}
\begin{figure*}
    \centering
    \includegraphics[width=0.9\textwidth]{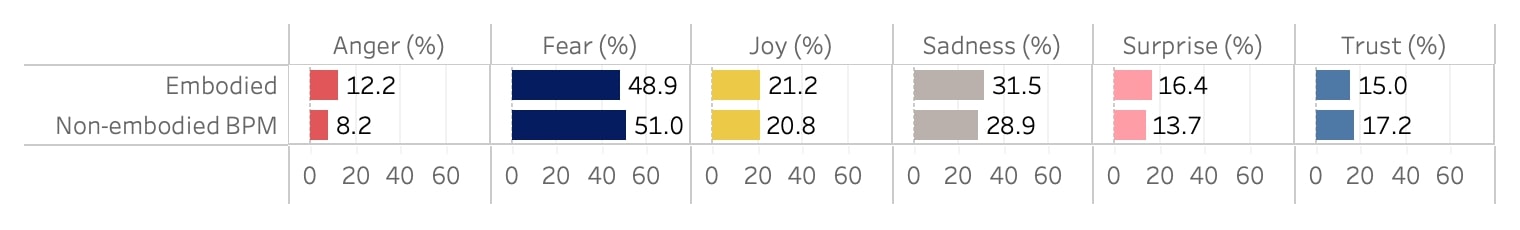}
    \vspace*{-4mm}
    \caption{BA2 - Spinn3r$_{\rm \it BPM-Zhuang}$ - 
    Percentages of embodied and non-embodied samples where the speaker is experiencing an emotion.}
    %Percentage of samples where the BPM possessor is annotated as experiencing each emotional category in the embodied vs. non-embodied samples (as annotated by \citeauthor{zhuang-etal-2024-heart}}
    \label{fig:spinn3r_embodied-annotations}
    \vspace*{-3mm}
\end{figure*}
\begin{figure*}
    \centering
    \includegraphics[width=0.9\textwidth]{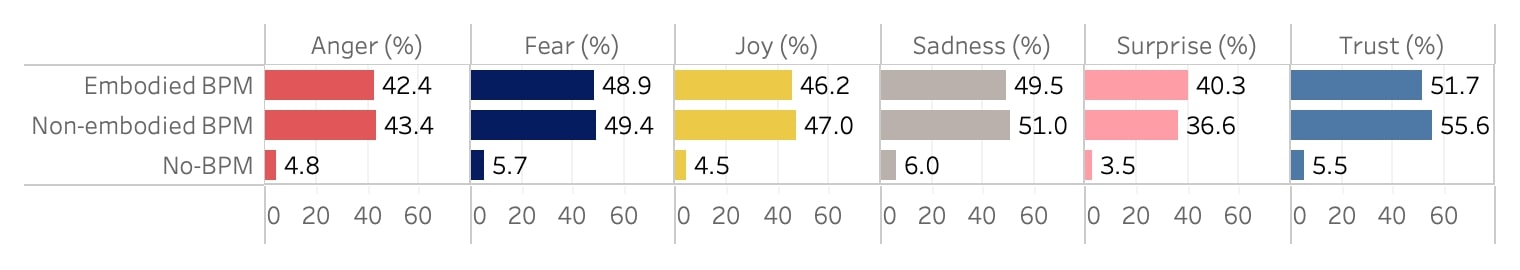}
    \vspace*{-4mm}
    \caption{BA2 - Spinn3r$_{\rm \it BPM-Zhuang}$ - 
    Percentages of embodied and non-embodied samples that include at least one word associated with an emotion.}
    % Percentage of samples including at least one word associated with each emotion in BPM samples annotated as embodied versus non-embodied (as annotated by \citeauthor{zhuang-etal-2024-heart}), as well as in all other samples containing no BPM.}
    \label{fig:spinn3r_embodied-lexicon-fig}
    \vspace*{-1mm}
\end{figure*}

\noindent \textit{Results:}  \Cref{fig:VAD_pronouns-fig.png} shows the results for valence, arousal and dominance.
We find that instances containing BPMs have higher percentages of emotion-associated co-occurring words than instances not containing BPMs.
\sw{This is true across 36/42 corpus--dimension pairs.}
% We find the same pattern for categorical emotions (\Cref{fig:emotion_pronouns-fig.jpg} in the Appendix), except for TUSC--Trust, where \textit{``my <BPM>''} instances have slightly lower scores than no BPM instances.
We also observe that the jump in scores from the no BPM corpus to BPM corpora is highest for the low-valence and low-dominance dimensions ($\sim$15 percentage points). \sw{We find that the BPM category has a statistically significant effect on the percentage of samples containing emotion-associated terms (more details on these tests can be found in \Cref{sec:emotion}}).  \\ 
%
%% Note that tweets are limited to 280 characters. However, blog posts can be longer. Tables \ref{table:BA2.spinn3r_VAD_bins} and \ref{table:BA2.spinn3r_emotion_bins} in the Appendix show the percentage of emotion-associated co-occurring words when controlling for blog post length for VAD and emotion categories, respectively.
%% Once again, we find the same trends discussed above.
%
%There are remarkable differences in the emotional signatures of different possessed BPMs---
% \textit{``my <BPM>''} instances co-occur markedly less with words associated with positive emotion, such as trust or joy or surprise, and they have lower co-occurrence with high valence and high dominance words than "your <BPM>" and "his/her/their <BPM>". \Cref{fig:tusc_ctry_radial} in \Cref{sec:emotion_pronouns} shows 
% \Cref{sec:emotion_pronouns} shows the proportions of tweets across all corpora.
% \Cref{sec:ba1.spinn3r} shows the full set of results. 
% \Cref{fig:tusc_city_radial} shows radial charts for TUSC$_{\rm \it city}$.
\noindent \textit{Discussion:} Referral to one's own body seems to display a strong co-occurrence with emotion-associated language, supporting theories of the connection between embodiment and emotion. When people discuss their own bodies, they tend to use more negative (low valence) emotional language and express less control (low dominance), suggesting these self-references often occur in contexts of pain or powerlessness. In Figure \ref{fig:wordcloud-top20-tusc-city} in the Appendix, we demonstrate that the most frequent words associated with BPMs in our corpora support this theory, such as \textit{hurt, sore,} and \textit{sick}.\\[3pt]
\noindent \textbf{BA2. What is the impact of explicitly embodied emotion on the emotions expressed through body part mentions?}\\
\noindent \textit{Method:} Body parts are often referenced as physically involved in emotional responses (e.g., \textit{my heart skipped a beat}, \textit{my stomach dropped}). \sw{Prior work (i.e. \citeauthor{zhuang-etal-2024-heart}) has assumed that such samples in natural language may be responsible for the emotional associations between emotional and body-related language.} In BA1, we showed that BPM instances are more likely to have emotion-associated co-terms than their no-BPM counterparts. %In 
\sm{With} this question we explore whether this increase only exists in explicitly embodied emotion. To do this, we analyze samples which are human-annotated by \citet{zhuang-etal-2024-heart} as either containing embodied emotion (where a body part physically participates in expressing the emotion---annotated as \textit{embodied}) or not (annotated as \textit{non-embodied}). 
Specifically, we looked at the degree of emotion-word co-occurrences in the embodied samples and in the non-embodied samples.
As a separate and complementary experiment,
we manually annotated these exact instances for whether the speaker was feeling any of the emotion categories (anger, fear, joy, sadness, surprise, and trust), by crowdsourcing on Amazon Mechanical Turk. We note a high inter-annotator agreement rate on our human annotations using Split-Half Class Match Percentage (90.1\% on 2 bins to 69.2\% on 10 bins). Further details on annotator agreement can be found in \Cref{sec:annotation}. % for all BPM possessors. 
This allows us to determine
% , in an arguably %a more direct way than lexical co-occurrences, 
\sm{the extent to which explicitly embodied BPMs are more or less emotional than 
% no-BPM instances and 
%the extent to which explicitly embodied BPMs are more or less emotional than 
not explicitly embodied BPM instances (non-embodied for short).}\\
%whether embodied BPM instances tended to be more emotional than non-embodied BPM instances.\\
%
%human readers perceive the emotional states inferred to be felt by BPM possessors as more emotionally intense depending on whether the body part mentioned is used in embodied emotion samples. Additionally, we can use the lexicons from prior experiments to see if emotion-associated co-terms are more likely to appear with embodied emotion samples. \\
\noindent \textit{Results:} \Cref{fig:spinn3r_embodied-annotations} shows 
the percentages of embodied and non-embodied samples where the speaker is experiencing an emotion.
% the percentage of instances where the BPM possessor is annotated as experiencing each emotional category, comparing 'Embodied BPM' vs 'Non-Embodied BPM' instances (as annotated by \citeauthor{zhuang-etal-2024-heart}. 
\Cref{fig:spinn3r_embodied-lexicon-fig} shows the percentages of embodied and non-embodied samples that include at least one word associated with an emotion. %"
%also compares these two groups by looking at the percentage of instances including at least one word associated with each emotional category, and also displaying the percentage for no-BPM instances in Spinn3r.
We find that both methods indicate no notable difference in the percentages of emotional samples across explicitly embodied versus non-embodied (or, more precisely, not \textit{explicitly} embodied) BPM instances
(<4 percentage points). 
% (<|4|\% difference between both groups using both methods, with a mix of positive and negative change). 
% Additionally, 
In contrast, there is a stark difference between the the emotion percentages of the no BPM samples and the embodied/non-embodied BPM samples ) (>30 percentage points).\\ % across all emotional categories).\\
\noindent \textit{Discussion:} 
The results show that emotional words appear more frequently in BPM sentences, regardless of whether or not the BPMs are explicitly written as physically connected to emotion in in the text (explicitly embodied). 
% BPM possessors 
Speakers are also equally likely to be expressing emotion whether the BPM is embodied or not. 
% Overall, this supports a deeper connection between BPMs and affect, independent of whether there is an explicit description of a physical role of of the BPM in the emotion. 
\sw{This unexpected finding supports the theory of constructed emotion and highlights a stronger connection between emotional expression and body-related language than theorized by previous works.}\\[3pt]
\noindent \textbf{BA3. Do individual body part mentions co-occur with markedly different emotion distributions?}\\[3pt]
\noindent \textit{Method:} 
%We are interested in finding whether different body parts are associated with different emotional contexts. 
In BA1, we looked at the co-occurrence of posts containing \textit{``my <BPM>''} with emotion-associated words. Here, we are interested in comparing this average score to posts containing specific \textit{``my <BPM>''} types. For the most common \textit{``my <BPM>''} types that 
% pass a threshold of minimum appearance of 
\sm{occur in at least}
100 instances across each corpus, we calculated the average proportion of %samples 
\sm{instances} that include words associated with specific emotion dimensions. We calculated a mean and standard deviation over these proportions and use these values to find \textit{``my <BPM>''} types which are significantly associated with particular emotional dimensions.
% For the most common \textit{``my <BPM>''} types (top 10 in Spinn3r$_{\rm \it BPM}$ and top 30 in TUSC$_{\rm \it city-BPM}$) across each corpus, we calculated the average proportion of samples that include words associated with specific emotion dimensions, and compare these values to overall proportions across all common \textit{``my <BPM>''} types identified. 
We also calculated the standard deviation for each emotional category across all common \textit{``my <BPM>''} types to identify which types are markedly associated with certain emotional dimensions.\\
% To investigate this, we use the same lexicons as in the previous experiment, and look at the percentage of noBPM samples with words contained as positively ranked for each emotional category as a baseline. Then, we evaluate the change in percentage of myBPM samples containing particular \textit{``my <BPM>''} types (i.e. "my arm", "my back") which contain words for each emotional category.
%
\begin{table*}[t]
\begin{small}
    \centering
    \resizebox{2.08\columnwidth}{!}{
    \begin{tabular}{lrrrrrrrr}
        \toprule
        & \multicolumn{2}{c}{Freq. Mental Distress} & \multicolumn{2}{c}{Freq. Phys. Distress} & \multicolumn{2}{c}{Life Expectancy} & \multicolumn{2}{c}{Physical Inactivity} \\
        \cmidrule(lr){2-3} \cmidrule(lr){4-5} \cmidrule(lr){6-7} \cmidrule(lr){8-9}
        & Spearman's \( r \) & \textit{p}-value & Spearman's \( r \) & \textit{p}-value & Spearman's \( r \) & \textit{p}-value & Spearman's \( r \) & \textit{p}-value \\
        \midrule
        a. Number of tweets & -0.170 & 0.418 & -0.167 & 0.425 & 0.290 & 0.160 & -0.243 & 0.242 \\
        b. Prop.\@ of \textit{<Fear word>} tweets & -0.230 & 0.231 & -0.370 & 0.054 & 0.160 & 0.403 & \textbf{-0.460} & \textbf{0.014} \\
        c. Prop.\@ of \textit{``my <BPM>''} tweets & \textbf{0.497} & \textbf{0.012} & \textbf{0.721} & \textbf{0.000} & \textbf{-0.409} & \textbf{0.043} & \textbf{0.704} & \textbf{0.000} \\
        d. Prop.\@ of \textit{"<BPM>"} tweets & \textbf{0.527} & \textbf{0.007} & \textbf{0.553} & \textbf{0.004} & \textbf{-0.613} & \textbf{0.001} & \textbf{0.539} & \textbf{0.006} \\
        \bottomrule
    \end{tabular}
    }
    \caption{Health Outcomes - TUSC$_{\rm \it city}$ - Spearman’s $r$ correlation %coefficients 
    and p-values showing the relationship between different health outcomes across cities and various features drawn from tweets from those cities.
    %the proportion of tweets containing the phrase \textit{``my <BPM>''} or \textit{<BPM>}, along with the total number of tweets in each city. \
    \textbf{Bolded} values indicate statistically significant correlations at $p < 0.05$.}
    \vspace*{-3mm}
    \label{tab:health_factors}
\end{small}
\end{table*}
\noindent \textit{Results:} We find that different \textit{``my <BPM>''} types are associated with different emotions to markedly different degrees, and that different profiles of associations for the same type can be found in different corpora. However, some \textit{``my <BPM>''} types carry consistent cross-corpus associations, such as \textit{my stomach} being most associated with sadness in both TUSC$_{\rm \it ctry-BPM}$ and Spinn3r$_{\rm \it BPM}$, whereas \textit{my chest} is most associated with anger. (Proportions for emotional word co-occurrence across \textit{``my <BPM>''} types for 
TUSC$_{\rm \it ctry-BPM}$ and Spinn3r$_{\rm \it BPM}$ are shown in Figures \ref{fig:tuscctry_vad-mybpms-fig} through  \ref{fig:spinn3r_vad-mybpms-fig} and the most associated emotion for each of the BPMs---which are often negative---are
%(i.e. anger, disgust, and fear, as opposed to joy), as we 
shown in \Cref{tab:top_emotions_for_mybpms} in the Appendix.)\\ % \Cref{sec:emotion}.\\
\textit{Discussion:} These results indicate that referral to one's body parts are associated with different affective expressions online. Overall trends in $TUSC_{city}$ seem to imply that referral to one's own body parts online often arise from situations of pain, lethargy, and a lack of control.

\section{Do BPMs Correlate with Health?}
The %trends we find across the 
previous two sections show that BPMs are common in online text and they exhibit many systematic and consistent trends across time and region, as well as w.r.t.\@ co-occurrence with emotion words.
% notably, the emotional signature of myBPMs tend toward negative and powerless emotions) 
These results are consistent with what we would expect if BPMs are linguistic indicators of one's health. In this section, we directly explore whether, at an aggregate level, the degree of BPMs in social media texts correlates with health outcomes.
% Although differences in health outcomes are difficult to measure for such large regions, our results indicate that regionally better health outcomes may be plausibly correlated to lower myBPM usage \cite{guyatt2007systematic, sandifer2021living}. 
We hypothesize that this occurs because BPMs are frequently used online by individuals to express pain or discomfort in their bodies. If so, regional discrepancies in BPM usage may also be correlated with different health outcomes.\\
% Based on our prior experiments (which indicate that emotional signature of BPMs in our corpora--especially myBPMs--tend toward negative and powerless emotions, as well as the regions we identify with higher BPM usage exhibiting poorer health outcomes), we wanted to find whether BPM usage may also be directly correlated to poorer health outcomes. 
\textit{Method:} To evaluate this hypothesis, we look at available city-wide health data \cite{CityHealthDashboard2025dataset} for all 25 American cities  in the TUSC$_{\rm \it city}$ dataset, and correlations %(Spearman's $r$ scores) 
between the proportion of regional tweets containing \textit{``my <BPM>''}/\textit{``BPM"} and four health measures: \textit{frequent mental distress, frequent physical distress, life expectancy, and physical inactivity}.\footnote{https://www.cityhealthdashboard.com}
As points of baseline comparison, we also look at how the health factors investigated are correlated with the number of tweets from each region,
and the correlation between the proportion of emotion-associated words (from the NRC Emotion/VAD lexicons) with the health outcomes studied.\\[3pt] 
%and the same four health outcomes.
% and find that there is no statistically signfificant relationship. 
\textit{Results:} 
% We find that regional BPM usage is highly correlated with negative health outcomes, as can be seen in 
\Cref{tab:health_factors} shows the Spearman rank correlations as well the p-values (we consider the correlations to be statistically significant if the p-value is below 0.05). Observe that the number of tweets per city is not correlated with the health outcomes (See Row A).
We find that most emotion--health outcome pairs are also not correlated or only slightly correlated. The highest correlation numbers are for fear--physical activity (See Row B).
(\Cref{tab:emotionwords-health-correlations} in the Appendix shows correlations for each of the emotion--health outcome pairs.) 
In contrast, the proportion of \textit{``<BPM>"} and \textit{``my <BPM>''} mentions (rows c and d) are moderately or strongly correlated with all three negative health outcomes and anticorrelated with life expectancy (statistically significant correlations are bolded).
Notably, frequent physical distress and physical inactivity are remarkably correlated with higher myBPM usage (Spearman's $r$ = 0.721, Spearman's $r$ = 0.704 respectively), and life expectancy is strongly negatively correlated with BPM use (Spearman's $r$ = -0.613). 
%We find statistically significant correlation between poorer health and proportion of BPM/myBPM tweets (relative to all tweets) on \textit{all four measures for both myBPM and anyBPM usage by city} ($p$-value < 0.05), which can be viewed in \Cref{tab:health_factors}. 
% As a baseline comparison, we also look at how this the health factors investigated are correlated with number of tweets from each region, and find that there is no statistically signfificant relationship. 
% between this baseline comparison and the health factors. 
Overall, these results show that simple metrics capturing the proportion of mentions of body parts in social media can be useful indicators of both physical and mental health. \sw{While our findings reveal a statistically significant correlation between body part mentions (BPMs) and regional health outcomes, we do not claim a direct causal relationship. It is likely that both language use and health indicators are shaped by broader social and demographic factors—such as educational attainment, economic status, or regional linguistic norms—which may contribute to the observed patterns. We see this possibility as an exciting result -- rather than treating BPMs as independent predictors of health, we interpret these correlations as evidence of shared variance that may offer insight into the sociolinguistic embedding of embodied experience. We highlight these associations as a starting point for future research that more directly models such confounding factors.}

\section{Conclusion}
% We define Body Part Mentions (BPMs) 
We created novel corpora designed specifically for the study of Body Part Mentions (BPMs), which includes the first-ever dataset of samples explicitly annotated for the emotions of human entities possessing BPMs.
Using these corpora, we answered a series of research questions on the significance of body-related words in everyday language, the relationship between embodiment and emotion, and factors correlated with BPM frequency such as emotional context, time of week/year, and region.
We showed that BPMs occur frequently in social media texts and have notable temporal and geographic trends.
We also showed that BPM instances have markedly higher emotion associations than non-BPM instances---with an especially marked increase in low valence (negativity) and low dominance (helplessness) instances.
Most notably, through experiments on data from 25 US cities, we showed that the degree of BPM usage can is a powerful indicator of aggregate-level well-being.
% Through our experiments, we find a likely connection %in our corpus 
% between speakers referring to their own body parts and expressions of pain, fatigue, and powerlessness. This hypothesis is supported by the strong correlation we find between the proportion of geo-tagged myBPMs tweeted from cities with poorer health outcomes. 
Although the connection between language, embodiment, and affect is now well-established, this paper is -- to our knowledge -- the first-ever approach to understanding this relationship grounded in large amounts of language data. We release our BPM corpus to the public, and hope that our work demonstrates body-related language as a rich and interesting source of material for future NLP research to investigate the deeper connection between language, embodiment, and emotional wellbeing. Although BPMs are a relatively simple tool for investigating the relationship between embodiment and everyday language, they offer a scalable, interpretable signal for understanding this connection empirically. We hope that our initial exploratory work, through showing that body-related language carries diverse and meaningful associations, emphasizes the richness of studying the intersection of embodiment and natural language.

% , and using computationally inexpensive and interpretable methods (such as BPM frequencies, co-occuring words, and pronouns preceding BPMs, emotion-association lexicons, and taking use of existing metadata in online corpora) 
% We showed that BPM analysis can easily produce interesting conclusions about the role of embodiment in everyday language. 

% Through future BPM analysis, NLP can offer a rich source of new hypotheses for researchers in affective sciences and public health. 
% a novel method which opens up many possibilities for future emotion annotation in NLP. 

\section*{Limitations}

Our work introduces the relevance of BPMs to NLP, and we argue for BPMs as a source of interesting research by demonstrating that their usage is correlated to the presence of emotional expression on social media as well as certain indicators of physical health and emotional wellbeing. But since we focus on BPMs occurring in a specific medium (online social media, specifically blog posts and tweets), much remains to be discovered about how body part words -- and their relationship to everyday language and affect -- manifest differently in other contexts.

Cultural and linguistic backgrounds significantly influence how people express emotions. Additionally, social media platforms and other digital communication channels produce unique language use patterns that may not reflect everyday language use in other environments (i.e., spoken conversation). We hope that in the future, other researchers can consider the relevance and limitations of producing BPM lists  and conducting similar experiments in other languages and with other datasets. This can both  extend our general knowledge of embodiment within language as well as help us consider the ways in which our results may differ in other linguistic contexts.

%\sw{
\sm{Since %our 
the primary aim of this paper is descriptive rather than explanatory to highlight that there are diverse and meaningful associations which BPMs carry, warranting future study, we do not isolate any specific mechanisms (i.e. effects of social variables, classification of BPM usage types, or deeper linguistic analysis) which could explain these associations. We encourage future work which can further probe the exact causes and explanations for the associations which we have discovered in this paper.

For this reason, we chose to use lexicon-based approaches to studying affect in our corpus, since although they cannot capture the full nuance of emotional expression in a single sample, especially in figurative or context-dependent language, they remain valuable tools for capturing broad trends across large corpora, and their interpretability makes them suitable for an exploratory, large-scale study like ours.  such as ML-based classifiers, could possibly help provide more fine-grained and context-sensitive emotion detection for more specific and contextual research questions (compared to the more general and exploratory research questions we attempt to answer in this paper).

}

\section*{Ethics Statement}

Our approach, as with any other data-driven approach to affective science/emotional wellbeing, should be considered an \textit{aggregate-level indicator} rather than a biomarker for individual's affective states \cite{guntuku2017detecting}. The measures we introduce for evaluating body part related words in everyday language, as well as their relationships to aspects of emotional and physical health should not be used as standalone indicators of these factors. Instead, they should be an additional metric that can be used in conjunction with a myriad of other investigative tools. This is especially important considering the diverse ways in which different individuals use words in everyday speech. Further best practices for ethical applications of emotional lexicons can be seen here: \cite{mohammad2022best}.

We also note that conceptions of emotion and wellbeing, especially as expressed through language, are heavily influenced by culture and linguistic variance \cite{barrett2008embodiment}. Interpretations of affective language may differ not only across languages but also within communities and individuals, shaped by socio-cultural norms, lived experiences, and context. As such, any claims or insights drawn from our analysis should be situated within a broader understanding of cultural and linguistic diversity, and we caution against universalizing interpretations without further cross-cultural validation.

\section*{Acknowledgments}
Many thanks to Tara Small for helpful discussions and comments. We also thank Lisa Pennel for her feedback and support on this paper.

\bibliography{anthology,custom}
\bibliographystyle{acl_natbib}

\appendix

\begin{table*}[t!]
\begin{small}
    \centering
\begin{tabular}{llp{8cm}r}
\toprule
\textbf{Dataset} & \textbf{Type} & \textbf{Description} & \textbf{\# Instances} \\ \midrule
1. Spinn3r & Blogs & English subset of ICWSM 2009 Spinn3r Blog Dataset. &80,379\\ 
2. Spinn3r$_{\rm \it BPM}$ & Blogs & Subset of Spinn3r containing only posts with at least one BPM  &8,371\\ 
3. Spinn3r$_{\rm \it noBPM}$ & Blogs & Subset of Spinn3r not containing any instances with a BPM  &72,008\\ 
4. Spinn3r$_{\rm \it myBPM}$   & Blogs & Subset of Spinn3r$_{\rm \it BPM}$ containing only instances including BPMs preceded by 'my'.& 1,391\\
6. Spinn3r$_{\rm \it yourBPM}$ & Blogs & Subset of Spinn3r$_{\rm \it BPM}$ containing only instances including BPMs preceded by 'your'.& 541\\
7. Spinn3r$_{\rm \it 3pBPM}$ & Blogs & Subset of Spinn3r$_{\rm \it BPM}$ containing only instances including BPMs preceded by 'his'/'her'/'their'.& 474\\

8. Spinn3r$_{\rm \it BPM-Zhuang}$ & Blogs  & Subset of Spinn3r$_{\rm \it BPM}$ where BPM mentions are annotated for embodied emotion by \cite{zhuang-etal-2024-heart}. & 6,359\\ 

% \hline
9. TUSC$_{\rm \it city}$ & Tweets & The TUSC$_{\rm \it city}$ dataset. &  104,575,991 \\ 
10. TUSC$_{\rm \it city-BPM}$ & Tweets & The TUSC$_{\rm \it city}$ dataset contatining only posts with at least one BPM. &  6,710,660 \\ 
11. TUSC$_{\rm \it city-myBPM}$   & Tweets & Subset of TUSC$_{\rm \it city-BPM}$ containing only instances including BPMs preceded by 'my'.& 1,060,507\\
12. TUSC$_{\rm \it city-yourBPM}$ & Tweets & Subset of TUSC$_{\rm \it city-BPM}$ containing only instances including BPMs preceded by 'your'.& 363,860\\
13. TUSC$_{\rm \it city-3pBPM}$ & Tweets & Subset of TUSC$_{\rm \it city-BPM}$ containing only instances including BPMs preceded by 'his'/'her'/'their'.& 338,510\\
14. TUSC$_{\rm \it ctry}$ & Tweets & The TUSC$_{\rm \it ctry}$ dataset. &  3,181,879 \\ 
15. TUSC$_{\rm \it ctry-BPM}$ & Tweets & The TUSC$_{\rm \it ctry}$ dataset contatining only posts with at least one BPM. & 231,577 \\ 
16. TUSC$_{\rm \it ctry-myBPM}$   & Tweets & Subset of TUSC$_{\rm \it ctry-BPM}$ containing only instances including BPMs preceded by 'my'.& 37,183\\
17. TUSC$_{\rm \it ctry-yourBPM}$ & Tweets & Subset of TUSC$_{\rm \it ctry-BPM}$ containing only instances including BPMs preceded by 'your'.& 12,936\\
18. TUSC$_{\rm \it ctry-3pBPM}$ & Tweets & Subset of TUSC$_{\rm \it ctry-BPM}$ containing only instances including BPMs preceded by 'his'/'her'/'their'.& 18,492\\ \bottomrule
\end{tabular}
\caption{Datasets used in this work.}
% TUSC$_{\rm \it city-BPM}$ and TUSC$_{\rm \it ctry-BPM}$ are released for public use (?), and Spinn3r datasets are available to approved researchers(?)
\label{tab:all_datasets}
\end{small}
\end{table*}

\section*{Appendix}

\section{Full Dataset Descriptions}
\label{sec:appendix_datasets}

All relevant datasets and subsets to this paper can be viewed in \Cref{tab:all_datasets}.

\section{Obtaining human ratings of emotion for Spinn3r$_{\rm \it BPM-Zhuang}$ corpus}
\label{sec:annotation}

The crowd-sourced annotations presented in this paper were approved by our Institutional Research Ethics Board. About 52\% of the annotators were male and about 48\% female, with average age of annotators being 39. Our final data collection process stored no information about annotator identity and as such there is no privacy risk to them. The annotators were free to do as many word annotations as they wished. The instructions included a brief description of the purpose of the task as well.

The key steps in producing the emotion annotation for this are:

\begin{enumerate}
    \setlength\itemsep{-0.5em}
    \item developing the questionnaire for emotion annotation
    \item developing measures for quality control (QC)
    \item annotating instances on the crowdsource platform (Amazon Mechanical Turk)
    \item discarding data from outlier annotations
    \item aggregating data from multiple annotators to determine final scores for each emotion
\end{enumerate}

We annotate the Spinn3r$_{\rm \it BPM-Zhuang}$ corpus, taken from \cite{zhuang-etal-2024-heart}, for the presence of six emotions. --  \textit{joy, fear/anxiety, sadness, anger, disgust, and trust} from Plutchik's wheel of basic emotions \cite{plutchik2001nature}.

\begin{figure*}
    \centering
    \includegraphics[width=\textwidth]{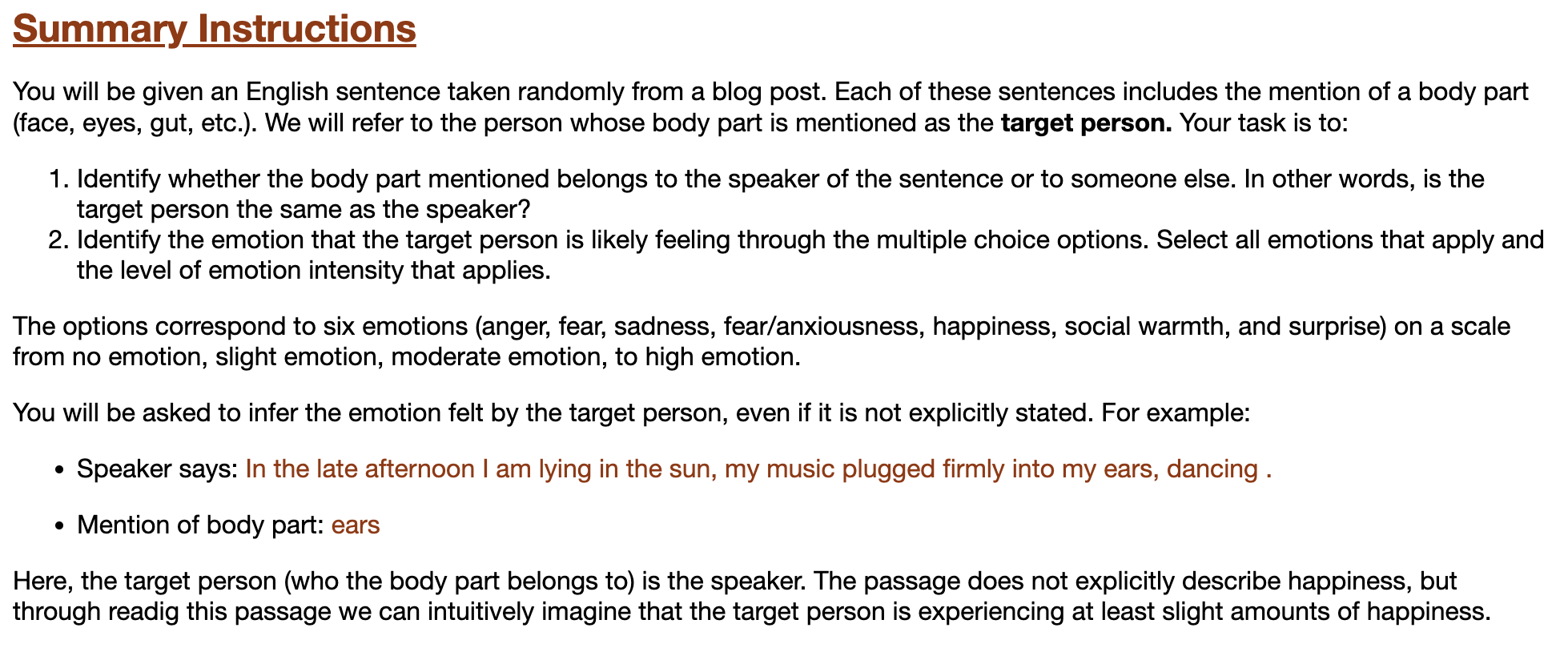}
    \caption{Summary instructions for crowdworkers annotating Spinner$_{BPM-Zhuang}$ on how to identify the body part `target person' and their emotion.}
    \label{fig:annot_1}
\end{figure*}

For each instance, we identify all possible BPMs. For each sample presented, we ask the crowdworker to identify whether a BPM in the sample belongs to the "speaker" or the "non-speaker". We also present them a description of the emotions we will want annotations for (See \Cref{fig:annot_1}). For example, in the sentence "Robin placed her hand on Kevin's shoulder", we would tell the annotator to identify the owner of "hand" or "shoulder", which would lead them to annotate for the emotion of Robin or Kevin respectively. We also note that all samples from the Spinn3r$_{\rm \it BPM-Zhuang}$ include BPMs that are preceded by the possessive pronoun "my", "his", or "her", guaranteeing that there is always an entity whose emotional state can be inferred from the BPM.

\begin{figure*}
    \centering
    \includegraphics[width=\textwidth]{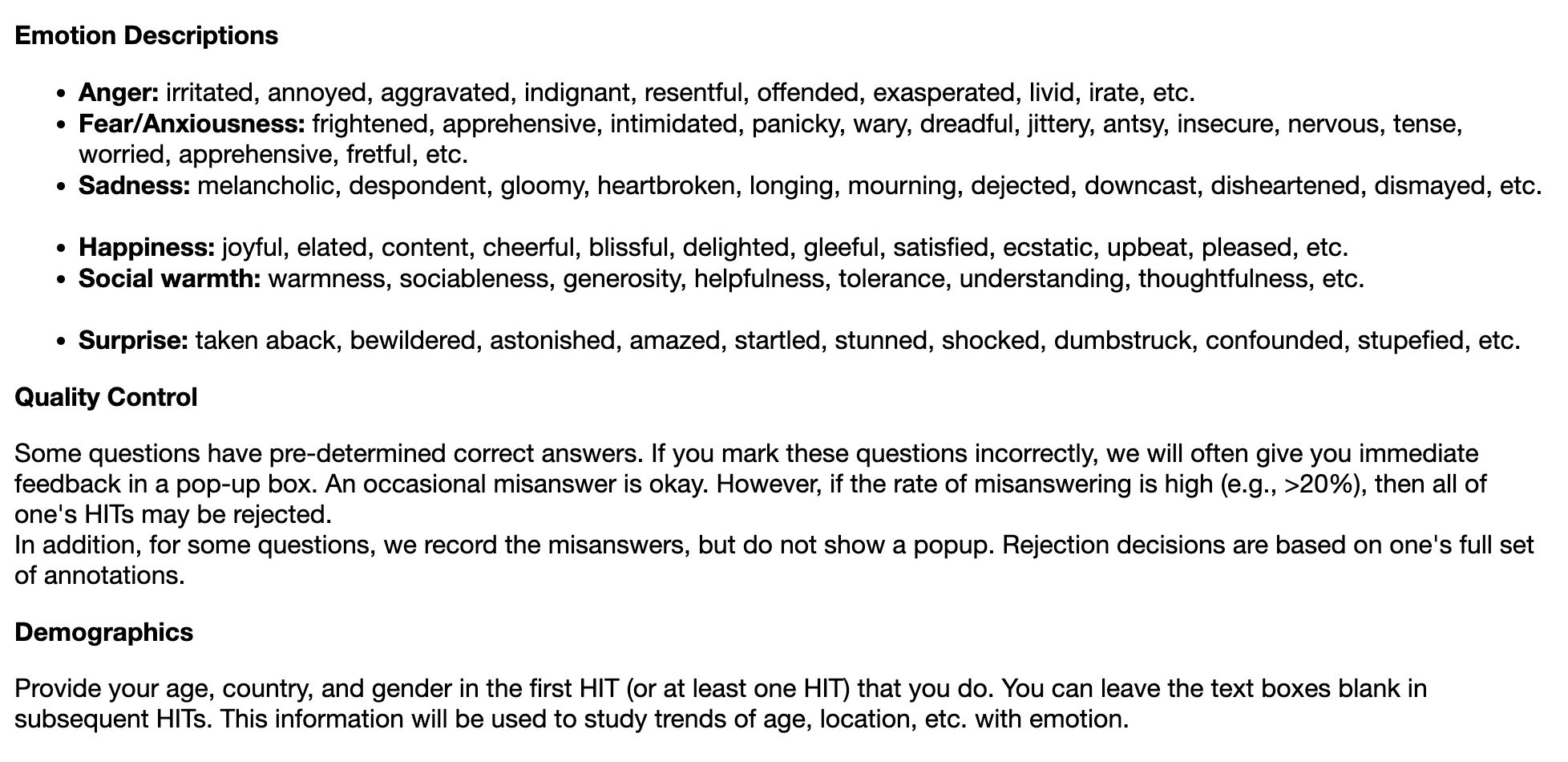}
    \caption{Instructions for crowdworkers annotating Spinner$_{BPM-Zhuang}$ on the various emotional categories to annotate.}
    %as well as contribute their demographic information.}
    \label{fig:annot_2}
\end{figure*}

\begin{figure*}
    \centering
    \includegraphics[width=\textwidth]{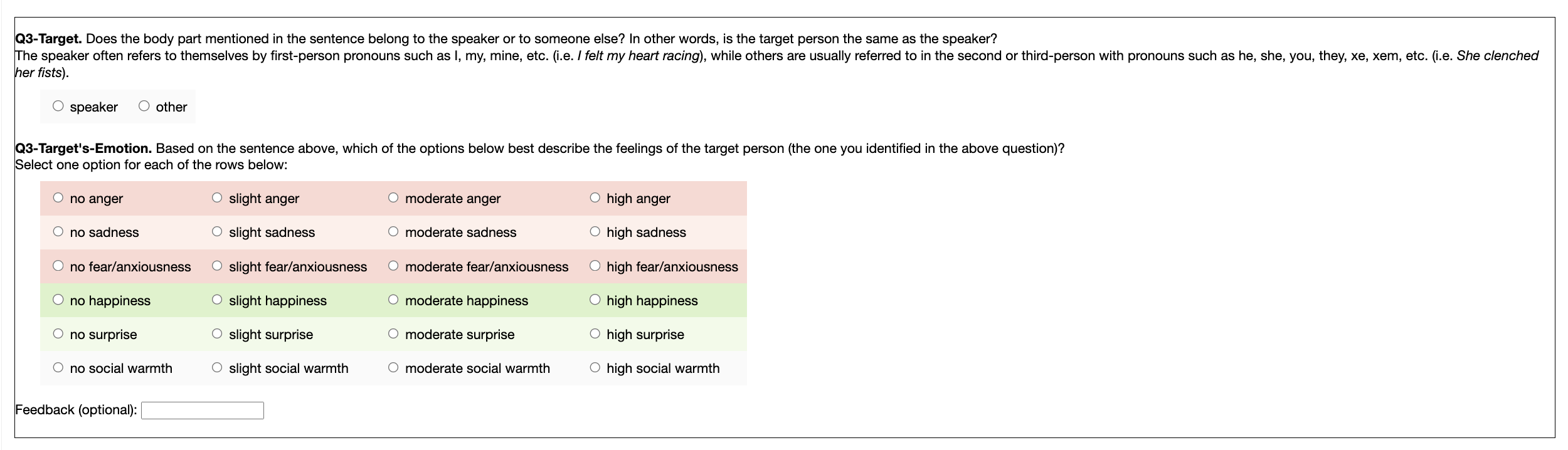}
    \caption{Questionnaire for annotating Spinner$_{BPM-Zhuang}$ for emotion felt by the BPM owner.}
    %where they can select the intensity of all emotions for the BPM owner, as well as identify if the BPM is owned by the 'speaker'.}
    \label{fig:annot_3}
\end{figure*}

We then present six emotional categories that they can annotate from, along with descriptions of these emotional categories (\Cref{fig:annot_2}). For each emotional category, they are five ranked categories they can choose from to indicate the severity of the emotion (no/slight/moderate/high/very high) (\Cref{fig:annot_3}).

Finally, we aggregate emotions to produce binary scores of an emotion being present/not present.

We assess annotation reliability using the Split-Half Class Match Percentage (SHCMP) as reported in the paper, a method adapted from traditional split-half reliability to handle categorical labels like those used for emotion intensity. SHCMP evaluates how consistently items are classified across multiple random groupings of the dataset. Specifically, the data is divided into n random subsets (with n = 2 representing a typical half-split) 1,000 times, and the average proportion of items that receive the same label across these splits is computed. A higher SHCMP score reflects greater reliability, indicating that the labels are likely to remain stable across repeated annotations.

\section{Most Frequent myBPMs by Corpus (Supplementary Table, B3)}
\label{sec:top_BPMs}

The top 20 most frequent myBPMs in each corpus, along with their frequency relative to all myBPMs present in their respective corpus, can be viewed in \Cref{tab:BPM_percentages_overall}.

% \clearpage
\begin{table*}[t]

\centering
{\small
\begin{tabular}{lrrr}
\toprule 
\textbf{BPM} & \textbf{Spinn3r$_{myBPM}$} & \textbf{TUSC$_{\rm \it ctry-myBPM}$} & \textbf{TUSC$_{\rm \it city-myBPM}$}\\
& \textit{\small{Blog sentences (\%)}}  & \textit{\small{Tweets (\%)}} & \small{\textit{Tweets (\%)}} \\ \midrule
my arm & - & 1.08 & 1.28 \\
my arms & 1.51 & - & - \\
my back & 1.88 & 4.61 & 3.95 \\
my blood & - & 1.14 & 1.07 \\
my body & 4.20 & 6.54 & 6.61 \\
my brain & 2.57 & 4.43 & 6.55 \\
my chest & 1.88 & 1.25 & 1.19 \\
my ears & 1.19 & - & - \\
my eye & 1.07 & - & - \\
my eyes & 10.23 & 1.40 & 1.21 \\
my face & 4.14 & 7.31 & 6.61 \\
my feet & 1.19 & 2.30 & 1.92 \\
my fingers & 1.07 & - & - \\
my hair & 3.26 & 11.18 & 10.08 \\
my hand & 2.82 & 2.21 & 2.07 \\
my hands & 3.32 & - & - \\
my head & 12.11 & 12.24 & 13.58 \\
my heart & 20.39 & 17.46 & 16.70 \\
my legs & - & - & - \\
my lips & - & - & - \\
my mouth & 1.82 & 2.89 & 2.94 \\
my neck & - & 1.34 & 1.29 \\
my nerves & - & - & 0.91 \\
my nose & - & 1.78 & 1.59 \\
my side & 1.51 & - & - \\
my skin & 2.13 & 1.83 & 1.87 \\
my stomach & 1.25 & 3.00 & 3.01 \\
my teeth & - & 1.21 & 1.13 \\
my throat & - & 1.13 & - \\
\midrule
\textbf{Total} & 79.54 & 86.36 & 85.58 \\
\bottomrule
\end{tabular}
}
\caption{B3 - Top 20 most common BPMs preceded by `my` throughout the Spinn3r$_{\rm \it BPM}$, TUSC$_{\rm \it ctry}$, and TUSC$_{\rm \it city}$ corpora with the frequency of appearance relative to total BPM distribution. The list shows the union of the top 20 unique BPMs for each dataset. Empty entry means that the BPM was not in the dataset's top 20.}
\label{tab:BPM_percentages_overall}
\end{table*}

% \begin{figure}
% \centering
% \includegraphics[width=0.47\textwidth]{figures/bpms_by_week_over_years.jpg}
% \caption{B4 - TUSC$_{ctry}$ - \% of tweets with at least one "my <BPM>" every weekday over each year from 2015 - 2021.}
% %where they can select the intensity of all emotions for the BPM owner, as well as identify if the BPM is owned by the 'speaker'.}
% \label{fig:bpms_by_week_over_years}

% \end{figure}

%\clearpage

% \section{Do individuals naturally use emotion-associated words when expressing their emotions through their body parts?}
% \label{sec:natural}

% \begin{figure}[t]
%     \centering
%     \includegraphics[width=\columnwidth]{figures/bpms_by_month_country.jpg}
%     \caption{B4 - TUSC$_{\rm \it ctry}$ - Percentage of tweets with at least one ``my <BPM>'' for different months. Bars are colored by season in northern hemisphere\footnote{\url{https://en.wikipedia.org/wiki/Season}} (winter: dark blue, spring: orange, summer: red, fall: brown).}
%     \label{fig:BPMs_by_month_tusc_country}
% \end{figure}

% \begin{figure}[t]
%     \centering
%     \includegraphics[width=\columnwidth]{figures/bpms_weekday.png}
%     \caption{B4 - TUSC$_{\rm \it ctry}$ - Percentage of tweets with at least one <BPM> for different weekdays.}
%     \label{fig:BPMs_by_weekday}
% \end{figure}

% \section{BPM type diversity}
\label{sec:bpm_type_diversity}

\begin{figure}[ht]
    \centering
    \includegraphics[width=0.9\columnwidth]{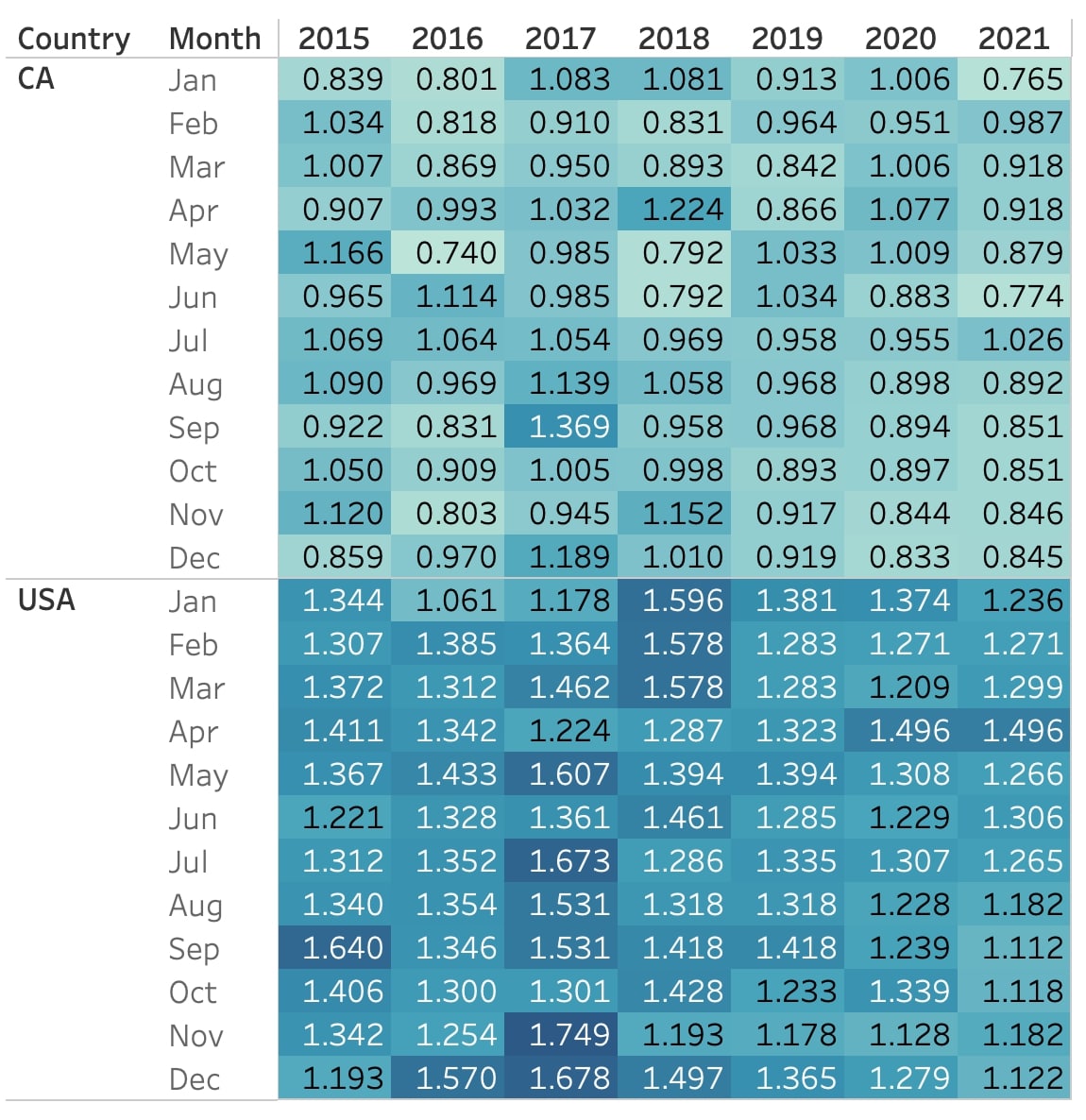}
    \caption{B5 - TUSC$_{\rm \it ctry}$ - \% of tweets with at least one ``my <BPM>`` for Canada and USA from 2015 to 2021 for each month.}
    \label{fig:BPMs_by_country_by_month}
\end{figure}

\section{BPM scores by region (Supplementary Figures and Statistical Tests, B5)}
\label{sec:bpm_ctry}

Figure \ref{fig:BPMs_by_country_by_month} displays the monthly percentage of tweets with at least one \textit{``my <BPM>''} from Canada and the USA between 2015 and 2021.

To assess whether the city of a post influences the probability of including a body-part mention (BPM), we modeled the number of posts containing BPMs out of the total posts per city using a binomial logistic regression, with city included as a categorical predictor. This approach models the log-odds of a post containing a BPM as a function of city. We then performed a likelihood ratio test comparing this model to a null model containing only an intercept, which indicated that cities overall have a highly significant effect on BPM usage (LR = 357.61, p < %10⁻⁶⁷
$10^{-67}$). When performing the same test using country as a categorical predictor, we find statistically significant results as well (LR = 9.8, p = 0.002), although its effect is considerably weaker than that of individual cities.

% \sw{

\section{Are mentions of the body associated with longer utterances: longer sentences/tweets?} % (Supplementary Experiment, BA1)}
\label{sec:longer}
% greater linguistic elaboration in online text?}\\[3pt]

% DO we want to include this???? If so will make this BA1 and modify the other numbers accordingly

\noindent \textit{Method:} We investigate whether BPMs are associated with more descriptive language by comparing samples with and without BPMs by length.\\
\noindent \textit{Results:} In Spinn3r, sentences containing BPMs are substantially longer than sentences without BPMs, with an average length of 482.07 characters compared to 146.53 characters for non-BPM sentences, making the average BPM sentence length 3.29x the length of the average non-BPM sentence. In the TUSC datasets, a similar trend holds but with a smaller magnitude: BPM tweets average 130.99 characters in TUSC${city}$ (vs.\@ 93.98 for non-BPM) and 133.42 characters in TUSC${ctry}$ (vs.\@ 101.45 for non-BPM), corresponding to a 1.39x and 1.32x increase, respectively.\\
\noindent \textit{Discussion:} These results suggest that body part mentions are consistently associated with longer sentences in a variety of online domains. The larger difference in the Spinn3r corpus may reflect the affordances of long-form narrative text, where BPMs may often occur in detailed narratives or reflective writing. 
Tweets often include just one sentence, but may at times include more; however, the total number of characters cannot exceed 280.
It is interesting that even such character-limited conditions,
tweets with BPMs are markedly longer than those without BPMs.
% More modest increases in tweets %length may be constrained by 
% is likely because of the 280 character Twitter character limits. Nonetheless, the consistent pattern across datasets indicates that references to the body may cue elaboration, possibly due to their centrality in conveying affective, sensory, or experiential content.\\

\section{Emotion associations between myBPMs, yourBPMs, 3pBPMs, noBPMs (Supplementary Figures, BA1)}
\label{sec:emotion_pronouns}

%\Cref{fig:tusc_ctry_radial} is a radial chart visualizing the ercentage of tweets with "my <BPM>", "your <BPM>", "his/her/their <BPM>", and "no BPM" that have at least one word associated with each emotional category according to the NRC Emotion/VAD lexicon.

% \Cref{fig:VAD_pronouns-fig.png} presents the percentage of sentences containing at least one high or low valence, arousal, or dominance word (NRC VAD lexicon) across different BPM categories (myBPM, yourBPM, 3pBPM, and noBPM) in each corpus.

\begin{figure*}
    \centering
    \includegraphics[width=\textwidth]{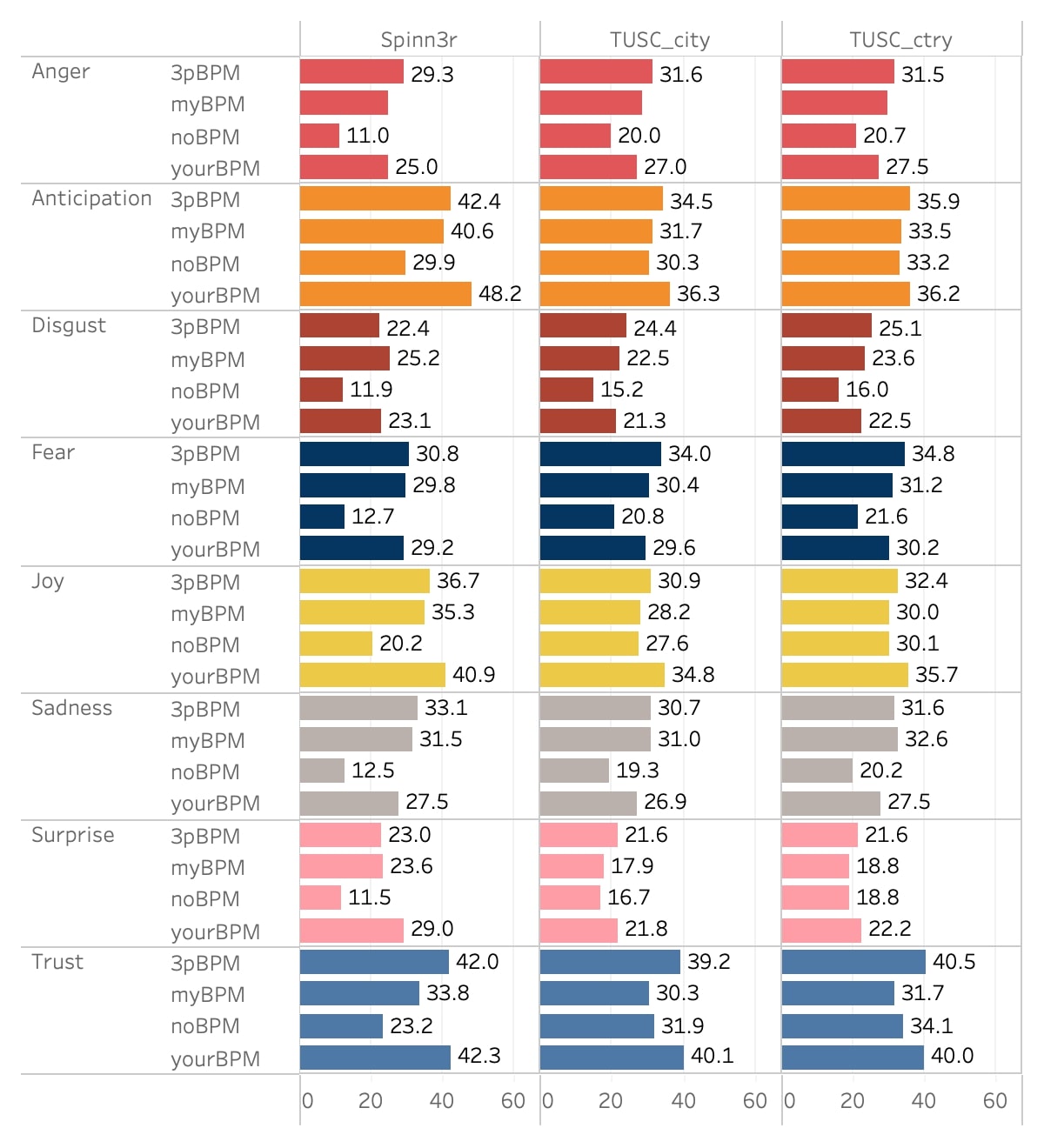}
    \caption{BA1 - Emotional categories. Percentage of sentences with at least one positive word in the eight emotions from Plutchik's emotion wheel (according to the NRC Emotion Lexicon) in each corpus in myBPM, yourBPM, 3pBPM, and noBPM categories.}
    \label{fig:emotion_pronouns-fig.jpg}
\end{figure*}

\Cref{fig:emotion_pronouns-fig.jpg} shows the percentage of sentences containing at least one positive word associated with Plutchik’s eight emotional categories (NRC Emotion Lexicon) across the same BPM categories in each corpus.

% \begin{figure}[t]
%     \centering
%     \includegraphics[width=\columnwidth]{figures/tusc_city_radial.png}
%     \caption{BA1:TUSC$_{\rm \it ctry-BPM}$ -
%     Radial charts displaying percentage of tweets with "my <BPM>", "your <BPM>", "his/her/their <BPM>", and "no BPM" that have at least one word associated with each emotional category according to the NRC Emotion/VAD lexicon.}
%     \label{fig:tusc_ctry_radial}
%     \vspace*{-5mm}
% \end{figure}

% \begin{figure*}
%     \centering
%     \includegraphics[width=0.85\textwidth]{figures/VAD_pronouns-fig.png}
%     \caption{BA1 - Valence, arousal, and dominance. Percentage of sentences with at least one high or low valence, arousal, or dominance word (according to the NRC VAD lexicon) in each corpus in myBPM, yourBPM, 3pBPM, and noBPM categories.}
%     \label{fig:VAD_pronouns-fig.png}
% \end{figure*}

% \section{Distribution of emotions in Spinn3r$_{embodied}$ according to our annotated data (Supplementary Figure, BA2)}
% \label{sec:embodied_annotation}

% Figure \ref{fig:spinn3r_embodied-annotations-fig} displays the proportion of sentences ranked under each emotion category using the annotated method, where the BPM possessor is explicitly annotated as experiencing the emotion in the BA2.Spinn3r$_{\rm \it BPM-Zhuang}$ dataset.

\section{ Controlling for Post Length (Supplementary Tables, BA2)}
\label{sec:BA2}
Note that tweets (which we use as individual posts for our tweet dataset, TUSC) are limited to 280 characters. However, blog sentences (which we use as individual posts for our blog dataset, Spinn3r) can be longer. Tables \ref{table:BA2.spinn3r_VAD_bins} and \ref{table:BA2.spinn3r_emotion_bins} in the Appendix show the percentage of samples co-occuring with emotion-associated words when controlling for blog post length for VAD and emotion categories, respectively.
%% Once again, we find the same trends discussed above.

\section{Effect of time on BPM usage (Supplementary statistical tests, B4)}
\label{sec:time_statisticaltests}

\sw{To assess the statistical significance of seasonal variation in BPM usage, we modeled the log-odds of a post containing a BPM as a function of month using a binomial logistic regression. We treat month as a continuous variable, starting from April, since this is the month where we can start to observe the trend of decreasing BPM use throughout the year. The results indicate a statistically significant decline in BPM usage over the months following April (coefficient = -0.0036, p < 0.001). When performing the same test with weekday, starting from Monday, we find a statistically significant decline in BPM usage from Monday as well (coefficient = -0.0104, p < 0.001).}

\section{Emotion associations between specific \textit{``my <BPM>''} types (Supplementary Figures and Statistical Tests, BA3)}
\label{sec:emotion}

% Figures \ref{fig:tuscctry_vad-mybpms-fig} and \ref{fig:tuscctry_emotion-mybpms-fig}  display the difference in the percentage of sentences containing emotion-associated words for samples containing "\textit{my <BPM>}",  for the top 30 most common body parts in TUSC$_{\rm \it city-BPM}$.

% Figures \ref{fig:spinn3r_vad-mybpms-fig} and \ref{fig:spinn3r_emotion-mybpms-fig} display the difference in percentage of sentences containing emotion-associated words for samples containing "\textit{my <BPM>}", for the top 15 most common body parts in TUSC$_{\rm \it city-BPM}$.

\begin{figure*}
    \centering
    \includegraphics[width=\textwidth]{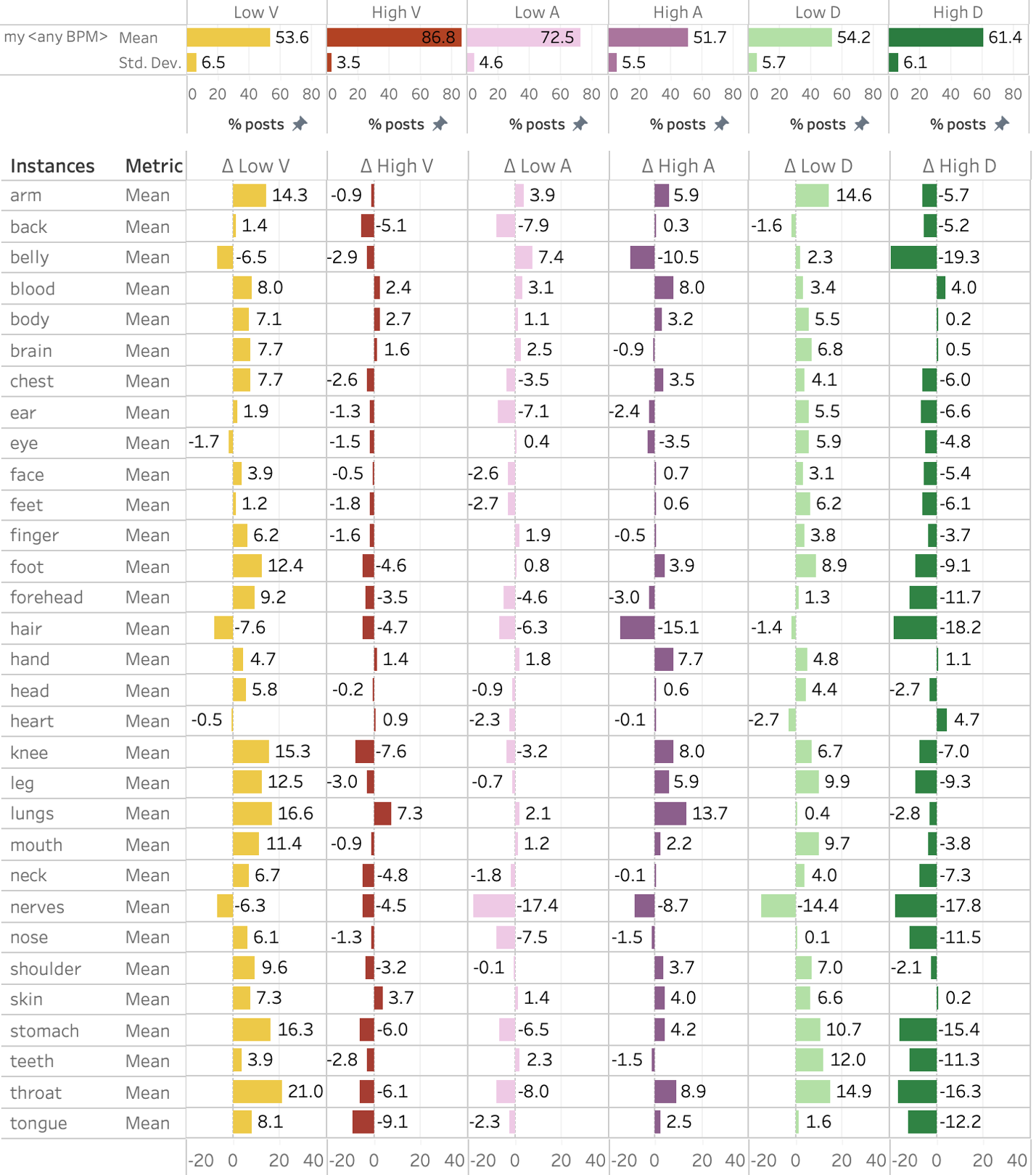}
    \caption{BA2 - TUSC$_{\rm \it ctry-myBPM}$ - Variance in emotion-associated term co-occurrence for top 30 most common \textit{``my <BPM>''} types present in the dataset. For each type, we display the delta in ("my <BPM>" type minus \textit{``my <BPM>''} mean) in the percentage of tweets with at least one word that is associated with high/low valence, arousal, and dominance (according to the NRC VAD lexicon). Mean and standard deviation are calculated over all body parts considered (top 30 most common \textit{``my <BPM>''} types present in the dataset).}
    \label{fig:tuscctry_vad-mybpms-fig}
\end{figure*}

\begin{figure*}
    \centering
    \includegraphics[width=\textwidth]{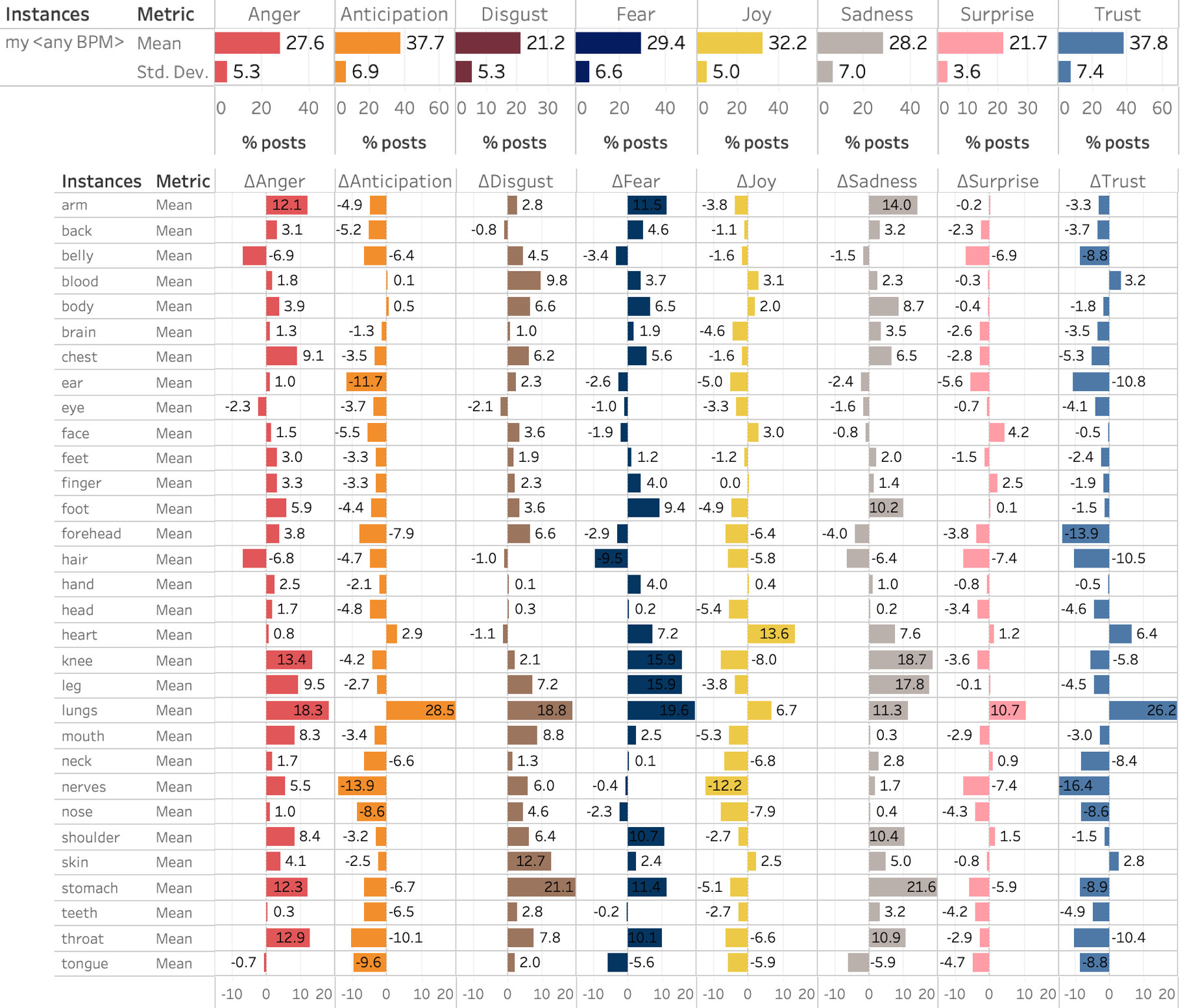}
    \caption{BA2 - TUSC$_{\rm \it ctry-myBPM}$ - Variance in emotion-associated term co-occurrence for top 30 most common \textit{``my <BPM>''} types present in the dataset. For each type, we display the delta in ("my <BPM>" type minus \textit{``my <BPM>''} mean) in the percentage of tweets with at least one word that is associated with each emotional category (according to the NRC Emotion Lexicon). Mean and standard deviation are calculated over all body parts considered (top 30 most common \textit{``my <BPM>''} types present in the dataset).}
    \label{fig:tuscctry_emotion-mybpms-fig}
\end{figure*}

\begin{figure*}
    \centering
    \includegraphics[width=\textwidth]{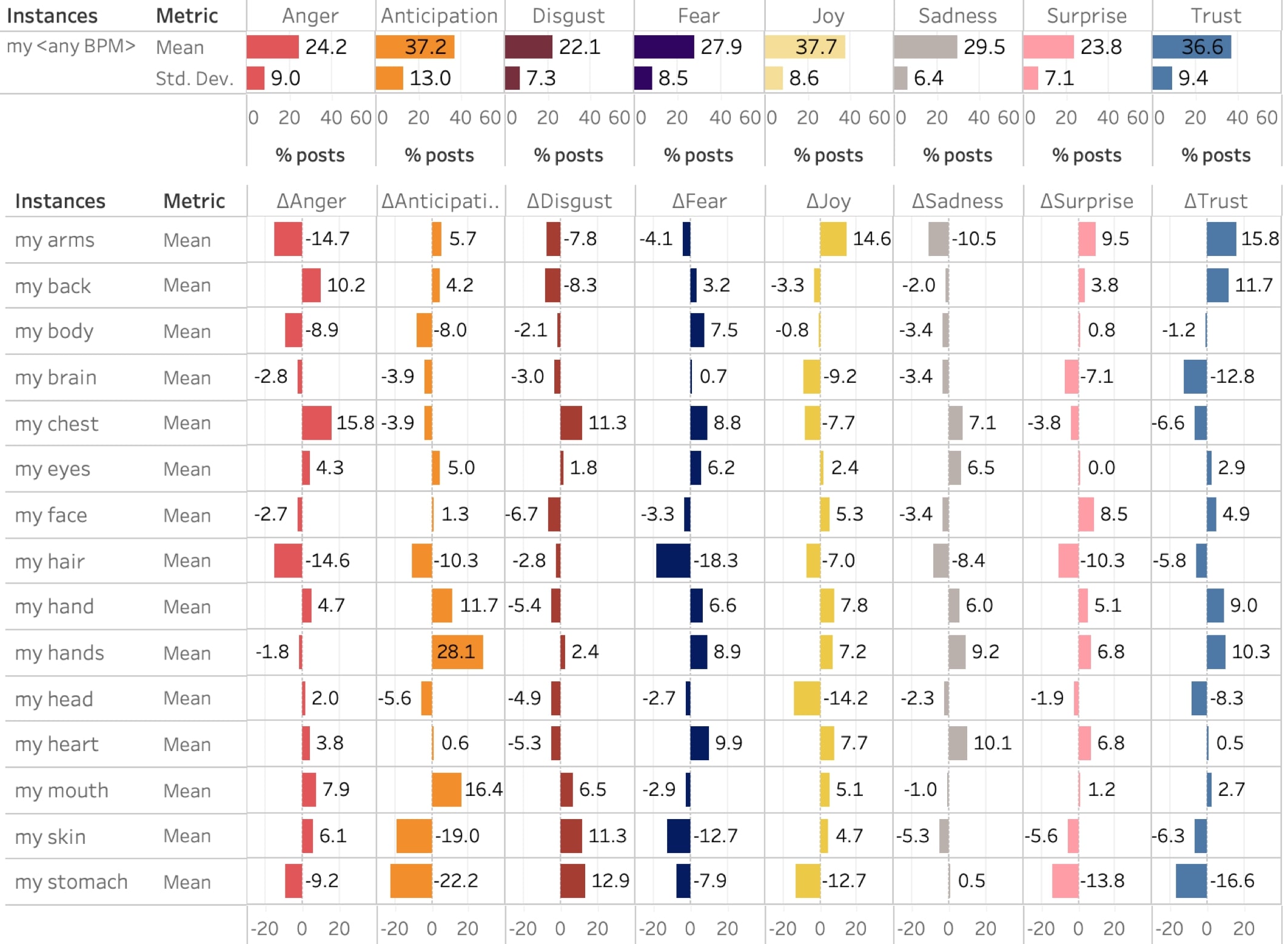}
    \caption{BA2 - Spinn3r$_{myBPM}$ - Variance in emotion-associated term co-occurrence for top 30 most common \textit{``my <BPM>''} types present in the dataset. For each type, we display the delta in ("my <BPM>" type minus \textit{``my <BPM>''} mean) in the percentage of blog sentences with at least one word that is associated with each emotional category (according to the NRC Emotion Lexicon). Mean and standard deviation are calculated over all body parts considered (top 30 most common \textit{``my <BPM>''} types present in the dataset).}
    \label{fig:spinn3r_emotion-mybpms-fig}
\end{figure*}

\begin{figure*}
    \centering
    \includegraphics[width=\textwidth]{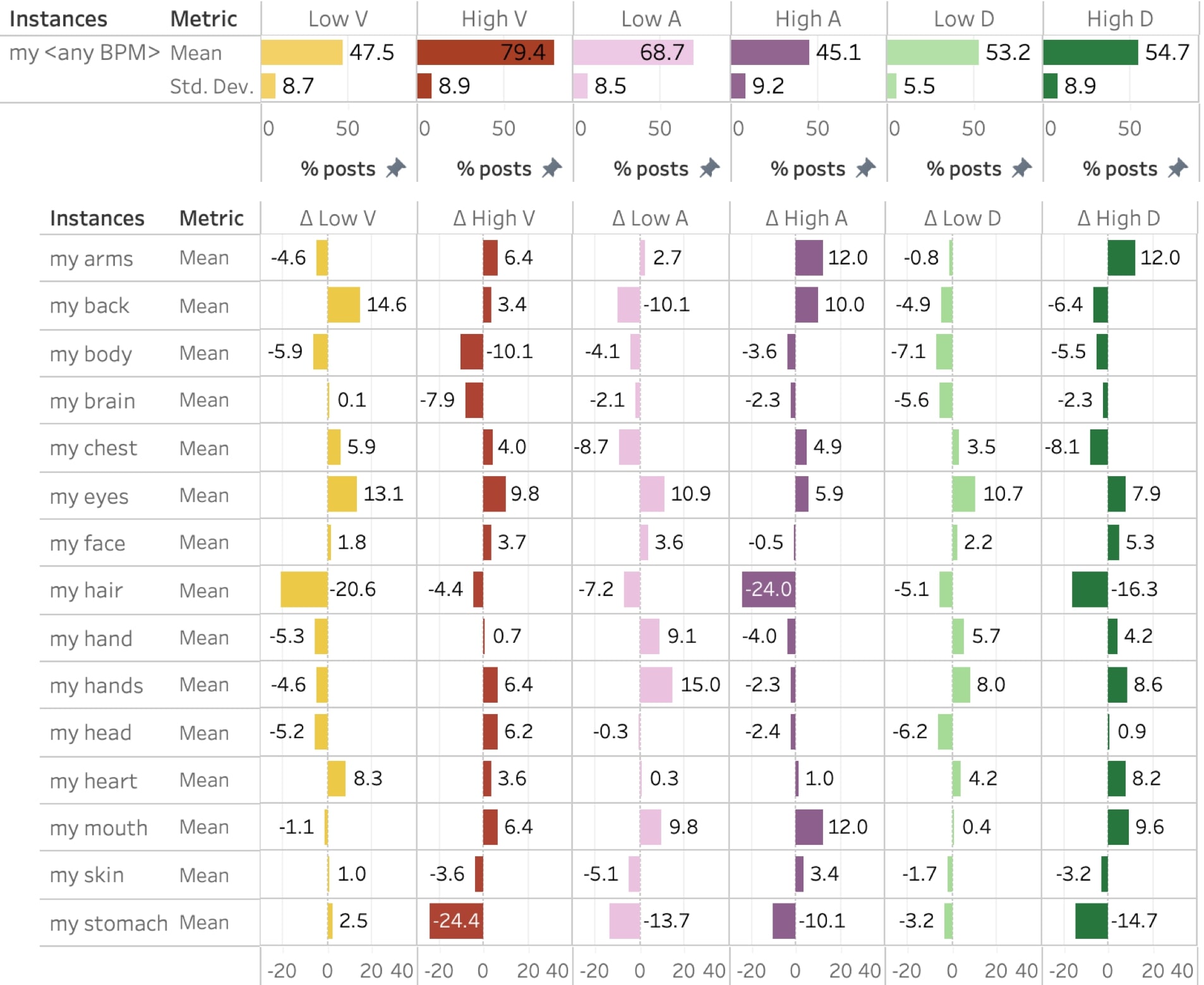}
    \caption{BA2 - Spinn3r$_{myBPM}$ - Variance in emotion-associated term co-occurrence for top 30 most common \textit{``my <BPM>''} types present in the dataset. For each type, we display the delta in ("my <BPM>" type minus \textit{``my <BPM>''} mean) in the percentage of blog sentences with at least one word that is associated with high/low valence, arousal, and dominance (according to the NRC VAD lexicon). Mean and standard deviation are calculated over all body parts considered (top 30 most common \textit{``my <BPM>''} types present in the dataset).}
    \label{fig:spinn3r_vad-mybpms-fig}
\end{figure*}

In this section, we include exact values for the differences in the percentage of sentences with emotion-associated words in samples containing "\textit{my <BPM>}" types in. These are shown in Figures \ref{fig:tuscctry_vad-mybpms-fig} (TUSC$_{\rm \it ctry-BPM}$ and VAD), \ref{fig:tuscctry_emotion-mybpms-fig} (TUSC$_{\rm \it ctry®-BPM}$ and emotion categories), \ref{fig:spinn3r_emotion-mybpms-fig} (Spinn3r$_{\rm \it BPM}$ and emotion categories), and  \ref{fig:spinn3r_vad-mybpms-fig} (Spinn3r$_{\rm \it BPM}$ and VAD). We focus on the top 30 most common \textit{``my <BPM>''} types in TUSC$_{\rm \it city-BPM}$. and the and top 15 most common body parts in Spinn3r$_{\rm \it BPM}$. The mean and standard deviation is also calculated over all common \textit{``my <BPM>''} types analyzed for each corpus for each emotional dimension, and for each \textit{``my <BPM>''} type we display the 'delta' as the proportion of the \textit{``my <BPM>''} type sample co-occuring with an emotion-associated words subtracted from the mean. All word-emotion associations are from the NRC VAD Lexicon and the NRC Emotion Lexicon.

\sw{We also use a two-way ANOVA test to examine the independent and interactive effects of BPM category and emotional category on VAD (Valence-Arousal-Dominance) values, which revealed significant main effects for both category (F(3,48) = 26.94, p < 0.001) and emotion (F(5,48) = 60.72, p < 0.001), with emotion showing the stronger effect, but no significant interaction between factors (F(15,48) = 1.25, p = 0.267). This indicates that while values vary across the different emotions (as expected, since some emotion-associated words naturally occur in certain corpora more frequently), the different BPM instance categories independently have a statistically significant effect emotion-associated word co-occurences. We also evaluate this test for the different emotional categories, again showing significant effects of the BPM categories (F(3,64) = 54.33, p < 0.001) and emotion (F(7,64) = 44.68, p < 0.001), and in this case also a significant interaction between category and emotion (F(21,64) = 2.03, p = 0.016) -- suggesting that the impact of BPM category is statistically significant, and also varies in strength depending on the specific emotion dimension.}

\section{How are body part words represented in word-emotion association lexicons? (Additional Experiment)}
\label{BPM-emotion-associations}
\noindent \textit{Method:} 144 BPMs from our list are found in the NRC Emotion Lexicon, and 200 are found in the arousal, dominance, and valence lexicon. All of the BPMs represented in the top 20 myBPMs across our corpora are represented in our lexicons (except for plural versions of the same BPM). We compare average scores for VAD and emotional categories using the NRC VAD lexicon and the NRC Emotion Lexicon respectively, for high frequency BPM words (defined as a word found in the top 20 myBPM list in any of our corpora), words that are in our BPM list, and words that are not in our BPM list.
\noindent \textit{Results:} We find that high frequency BPMs exhibit high changes in associations with valence, arousal, and dominance from non-BPMs (significantly  than the average BPM score). However, frequent BPMs are rarely ranked as positive instances for any of the seven emotional categories (less than the non-BPM baseline for all categories except for Surprise and Disgust).
\noindent \textit{Discussion:} Although common BPMs seem to have a different emotional signature than non-BPM words, they seem to have little everyday association to particular emotional categories. This corroborates the theory that bodily interpretations form the basis of our most basic emotional categories, but that more specific emotional categories are produced from contextual interpretations of these bodily signals.

\section{How does referring to one's own body change the emotional signature in personal narratives online (BA1: Supplementary Figures)}

\label{sec:ba1.spinn3r}

\Cref{table:BA2.spinn3r_VAD_bins} displays the percentage of BPM vs no BPM sentences across High/Low VAD categories containing at least one word associated with each emotional category according to NRC VAD lexicon in the Spinn3r dataset across different bins of sentence length for high/low VAD, and \Cref{table:BA2.spinn3r_emotion_bins} displays this data for emotional categories. 

\begin{table*}[t]
  \centering
  \caption{BA2 - Spinn3r - Percentage of BPM vs no BPM sentences across High/Low VAD categories containing at least one word associated with each emotional category according to NRC VAD lexicon in the Spinn3r dataset.}
  \resizebox{2.08\columnwidth}{!}{
    \begin{tabular}{lcccccccccccc}
        \hline
        Bin & \multicolumn{2}{c}{High Valence} & \multicolumn{2}{c}{Low Valence} & \multicolumn{2}{c}{High Arousal} & \multicolumn{2}{c}{Low Arousal} & \multicolumn{2}{c}{High Dominance} & \multicolumn{2}{c}{Low Dominance} \\
        (\# of words) & \textit{``my <BPM>''} & no BPM & \textit{``my <BPM>''} & no BPM & \textit{``my <BPM>''} & no BPM & \textit{``my <BPM>''} & no BPM & \textit{``my <BPM>''} & no BPM & \textit{``my <BPM>''} & no BPM \\
        \hline
        (0,10] & 0.38 & 0.40 & 0.19 & 0.12 & 0.14 & 0.11 & 0.27 & 0.24 & 0.15 & 0.20 & 0.17 & 0.13 \\
        (10,20] & 0.77 & 0.80 & 0.39 & 0.28 & 0.32 & 0.29 & 0.57 & 0.60 & 0.36 & 0.50 & 0.37 & 0.33 \\
        (20,30] & 0.89 & 0.93 & 0.47 & 0.40 & 0.44 & 0.44 & 0.78 & 0.79 & 0.58 & 0.69 & 0.62 & 0.47 \\
        (30,40] & 0.98 & 0.97 & 0.66 & 0.53 & 0.55 & 0.56 & 0.92 & 0.88 & 0.73 & 0.79 & 0.71 & 0.60 \\
        (40,50] & 0.98 & 0.97 & 0.69 & 0.57 & 0.65 & 0.60 & 0.91 & 0.90 & 0.83 & 0.83 & 0.79 & 0.65 \\
        \hline \\
    \end{tabular}
  }
  \label{table:BA2.spinn3r_VAD_bins}
\end{table*}

\begin{table*}[t]
  \centering
    \caption{Percentage of BPM vs no BPM sentences across bins containing at least one word associated with each emotional category (anger, fear, joy, sadness, surprise, trust) according to the NRC VAD Lexicon in Spinn3r.}
   \resizebox{2.08\columnwidth}{!}{
    \begin{tabular}{lcccccccccccc}
        \hline
        Bin 
        & \multicolumn{2}{c}{Anger} & \multicolumn{2}{c}{Fear} & \multicolumn{2}{c}{Joy} & \multicolumn{2}{c}{Sadness} & \multicolumn{2}{c}{Surprise} & \multicolumn{2}{c}{Trust} \\
        (\# of words) & \textit{``my <BPM>''} & no BPM & \textit{``my <BPM>''} & no BPM & \textit{``my <BPM>''} & no BPM & \textit{``my <BPM>''} & no BPM & \textit{``my <BPM>''} & no BPM & \textit{``my <BPM>''} & no BPM\\
        \hline
        (0,10] & 0.07 & 0.04 & 0.10 & 0.05 & 0.11 & 0.09 & 0.10 & 0.05 & 0.06 & 0.05 & 0.09 & 0.10 \\
        (10,20] & 0.20 & 0.11 & 0.21 & 0.13 & 0.19 & 0.22 & 0.24 & 0.13 & 0.12 & 0.12 & 0.18 & 0.25 \\
        (20,30] & 0.19 & 0.18 & 0.24 & 0.22 & 0.32 & 0.32 & 0.27 & 0.21 & 0.23 & 0.19 & 0.35 & 0.39 \\
        (30,40] & 0.30 & 0.26 & 0.39 & 0.30 & 0.50 & 0.44 & 0.40 & 0.29 & 0.28 & 0.27 & 0.46 & 0.51 \\
        (40,50] & 0.44 & 0.31 & 0.51 & 0.33 & 0.56 & 0.48 & 0.48 & 0.33 & 0.44 & 0.29 & 0.52 & 0.54 \\
        (50,60] & 0.36 & 0.19 & 0.41 & 0.23 & 0.68 & 0.41 & 0.36 & 0.21 & 0.45 & 0.26 & 0.68 & 0.48 \\
        \hline
    \end{tabular}
    }
    \label{table:BA2.spinn3r_emotion_bins}
\end{table*}

\section{What are the emotions most commonly associated with the most frequently discussed body parts?}
\label{sec:top_emotions_for_bpms}

In \Cref{tab:top_emotions_for_mybpms}, we display the top emotion (with associated increase in emotion-associated word from \textit{``my <BPM>''} average) for top \textit{``my <BPM>''} types across Spinn3r$_{BPM}$ and TUSC$_{BPM}$ datasets. A dash indicates the body part is not present in top \textit{``my <BPM>''} types considered for the dataset (top 15 for Spinn3r$_{BPM}$ and top 30 for TUSC$_{BPM}$).

\begin{table*}[htbp]
\centering
\small
\begin{tabular}{lll}
\toprule
\textbf{Body Part} & \textbf{Spinn3r (Top Emotion)} & \textbf{TUSC (Top Emotion)} \\
\midrule
arms & Trust (15.8) & - \\
arm & - & Anger (12.1) \\
back & Trust (11.7) & Fear (4.6) \\
belly & - & Disust (4.5) \\
blood & - & Disgust (9.8) \\
body & Surprise (0.8) & Disgust (6.6) \\
brain & Fear (0.7) & Sadness (3.5) \\
chest & Anger (15.8) & Anger (9.1) \\
ear & - & Disgust (2.3) \\
eye & - & Surprise (-0.7) \\
face & Surprise (8.5) & Surprise (4.2) \\
feet & - & Anger (3.0)  \\
finger & - & Fear (4.0) \\
foot & - & Sadness (10.2) \\
forehead & - & Disgust (6.6) \\
hair & Disgust (-2.8) & Disgust (-1.0) \\
hand & Anticipation (11.7) & Fear (4.0) \\
hands & Anticipation (28.1) & Anger (1.7) \\
head & Anger (2.0) & Disgust (0.3) \\
heart & Sadness (10.1) & Joy (13.6) \\
knee & - & Sadness (18.7) \\
leg & - & Sadness (17.8) \\
lungs & - & Anticipation (28.5) \\
mouth & Anticipation (16.4) & Disgust (8.8) \\
neck & - & Sadness (2.8) \\
nerves & - & Disgust (6.0) \\
nose & - & Disgust (4.6) \\
shoulder & - & Anger (8.4) \\
skin & Disgust (11.7) & Disgust (12.7) \\
stomach & Disgust (12.9) & Sadness (21.6) \\
teeth & - & Sadness (3.2) \\
throat & - & Anger (12.9) \\
tongue & - & Disgust (2.0) \\
\bottomrule
\end{tabular}
\caption{Most associated emotion (with associated increase in emotion-associated word from \textit{``my <BPM>''} average) for  \textit{``my <BPM>''} types across Spinn3r$_{BPM}$ and TUSC$_{BPM}$ datasets (top 15 for Spinn3r$_{BPM}$ and top 30 for TUSC$_{BPM}$). A dash indicates the body part is not present in top \textit{``my <BPM>''} types considered for the dataset.}
\label{tab:top_emotions_for_mybpms}
\end{table*}

\section{Is physical wellbeing correlated with emotional word use? (Additional Experiment)} 
\label{sec:emotionwords-health-correlation}

We also look at whether the physical wellbeing indicators we examine in other experiments are correlated with emotion-related words according to the NRC lexicon. See \Cref{tab:emotionwords-health-correlations}.

\begin{table*}[htbp]
\centering
\small
\begin{tabular}{lcccc}
\toprule
\textbf{Emotional Category} & \textbf{Mental Distress} & \textbf{Physical Distress} & \textbf{Life Expectancy} & \textbf{Physical Inactivity} \\
\midrule
Anger & -0.05 (p=0.8170) & -0.12 (p=0.5520) & -0.16 (p=0.4030) & -0.12 (p=0.5480) \\
Anticipation & -0.10 (p=0.6110) & -0.24 (p=0.2090) & 0.07 (p=0.7260) & -0.33 (p=0.0870) \\
Disgust & 0.07 (p=0.7090) & 0.07 (p=0.7200) & -0.23 (p=0.2340) & 0.09 (p=0.6610) \\
Fear & -0.23 (p=0.2310) & -0.37 (p=0.0540) & 0.16 (p=0.4030) & \textbf{-0.46 (p=0.0140)} \\
High Arousal & -0.12 (p=0.5350) & -0.25 (p=0.2070) & 0.03 (p=0.8770) & -0.34 (p=0.0760) \\
High Dominance & -0.18 (p=0.3570) & -0.31 (p=0.1090) & 0.16 (p=0.4210) & \textbf{-0.38 (p=0.0440)} \\
High Valence & -0.13 (p=0.4950) & -0.24 (p=0.2110) & 0.12 (p=0.5390) & -0.33 (p=0.0900) \\
Joy & -0.05 (p=0.7960) & -0.15 (p=0.4580) & 0.09 (p=0.6490) & -0.24 (p=0.2260) \\
Low Arousal & -0.11 (p=0.5710) & -0.27 (p=0.1590) & 0.05 (p=0.8090) & -0.35 (p=0.0700) \\
Low Dominance & -0.13 (p=0.5050) & -0.26 (p=0.1830) & 0.06 (p=0.7710) & -0.37 (p=0.0530) \\
Low Valence & -0.19 (p=0.3400) & -0.26 (p=0.1790) & 0.07 (p=0.7110) & \textbf{-0.38 (p=0.0460)} \\
Sadness & -0.15 (p=0.4430) & -0.27 (p=0.1600) & 0.08 (p=0.6710) & \textbf{-0.39 (p=0.0410)} \\
Surprise & -0.10 (p=0.6300) & -0.23 (p=0.2490) & 0.09 (p=0.6380) & -0.33 (p=0.0820) \\
Trust & -0.16 (p=0.4050) & -0.31 (p=0.1120) & 0.13 (p=0.5190) & \textbf{-0.38 (p=0.0480)} \\
\bottomrule
\end{tabular}
\caption{Spearman's $\rho$ and p-values between proportion of emotional words and city-level health outcomes. Bolded values are statistically significant at $p<0.05$.}
\label{tab:emotionwords-health-correlations}
\end{table*}

\section{Why/when do we refer to our own bodies? (Additional Experiment)}
\label{appendix:why_when_BPMs}

\begin{figure*}
    \centering
    \includegraphics[width=\textwidth]{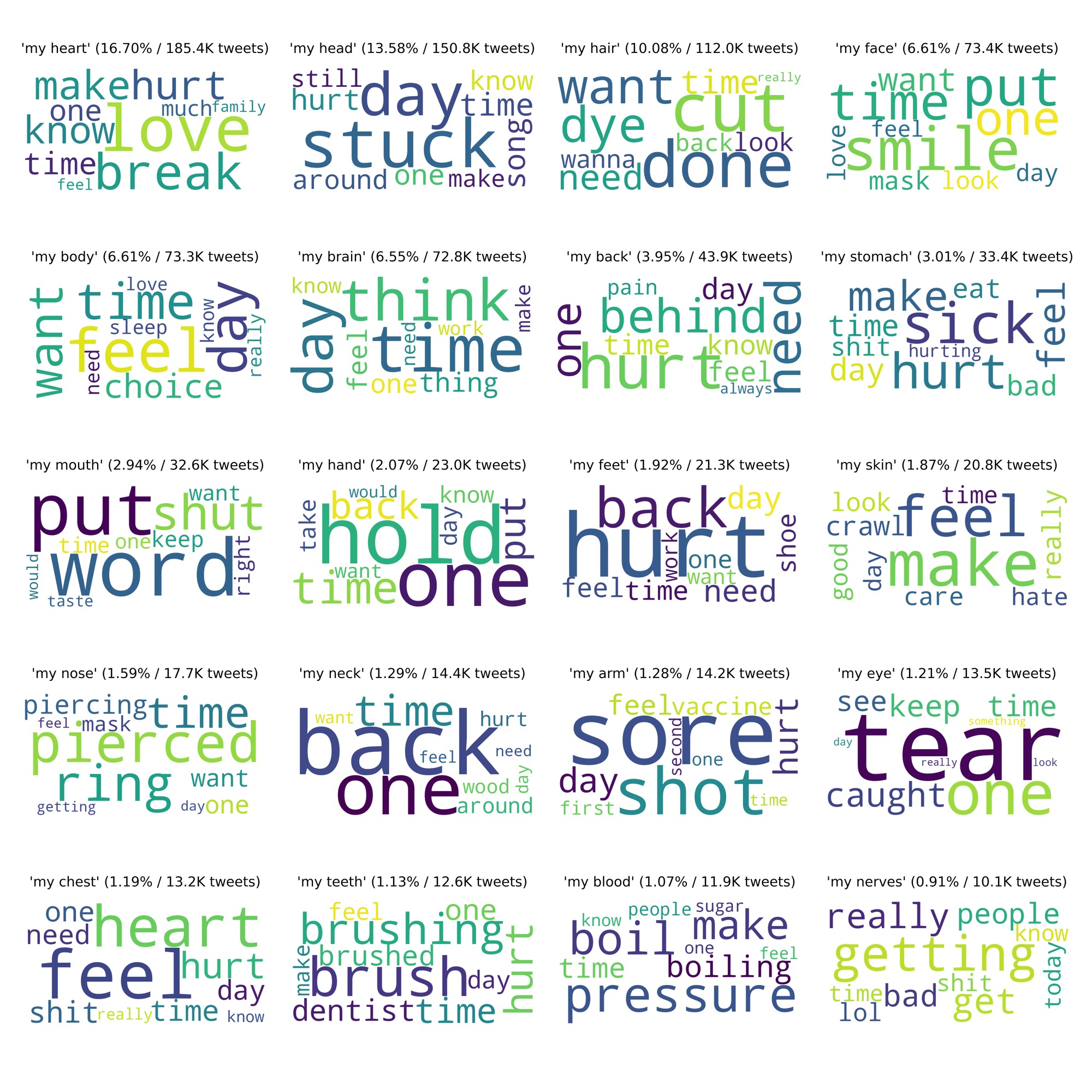}
    \caption{B6 - Wordclouds for the twenty most frequent \textit{``my <BPM>''} types in the TUSC$_{\rm \it city}$ dataset with the most frequent co-occurring words.}
    \label{fig:wordcloud-top20-tusc-city}
\end{figure*}

\noindent \textit{Method:} We evaluate the context of words that tend to surround myBPMs by looking at word clouds which visualize the words which are most likely to appear within the context window of particular myBPMs (See \Cref{fig:wordcloud-top20-tusc-city}).\\
\noindent \textit{Results:} We find that there are significantly more 3pBPM types with $>0.1\%$ occurrence compared to \textit{``my <BPM>''} types with $>0.1\%$  (131 vs 57 in the Spinn3r$_{\rm \it BPM}$ dataset, and 108 vs 56 in the TUSC$_{\rm \it ctry-BPM}$ dataset). We also find that, although \textit{``my <BPM>''} types exhibit a rich diversity in associated contexts, that some \textit{``my <BPM>''} types share common contexts as well, especially "hurt", "pain", and "sick", which frequently co-occur with several frequent myBPMs such as "my head", "my back", "my neck", and" my stomach".\\
\noindent \textit{Discussion:} The analysis reveals that third-person BPM types (3pBPM) in are significantly more diverse than \textit{``my <BPM>''} types at the 0.1\% occurrence threshold in the Spinn3r dataset, indicating a more limited and concentrated vocabulary when people refer to their own body than the bodies of others. The words with negative associations with health frequently accompanying some of the most common \textit{``my <BPM>''} types also highlight health concerns and physical pain as central themes for myBPM usage.

% \clearpage

\end{document}